\documentclass[lettersize,journal]{IEEEtran}
\usepackage{amsmath,amsfonts}
\usepackage{algorithmic}
\usepackage{algorithm}
\usepackage{array}
\usepackage[caption=false,font=footnotesize,labelfont=sf,textfont=sf]{subfig}
\usepackage{textcomp}
\usepackage{stfloats}
\usepackage{url}
\usepackage{verbatim}
\usepackage{graphicx}
\usepackage{cite}
\usepackage[table,xcdraw]{xcolor}
\usepackage{tabularx}
\usepackage{colortbl}
\usepackage{adjustbox}
\usepackage{multirow} 
\usepackage{booktabs}
\usepackage[citecolor=blue, colorlinks]{hyperref}
\newcommand{\ADD}[1]{\textcolor{black}{\textbf{}#1}}
\newcommand{\ADDCZ}[1]{\textcolor{black}{\textbf{}#1}}
\newcommand{\ADDCZAQ}[1]{\textcolor{black}{\textbf{}#1}}

\begin{document}
\title{Objective Quality Assessment of Point Clouds Using Multi-scale Implicit Structural Similarity}
\author{Zhang Chen, Shuai Wan,~\IEEEmembership{Member,~IEEE,} Yuezhe Zhang, Siyu Ren, Fuzheng Yang,~\IEEEmembership{Member,~IEEE}, \\ and Junhui Hou,~\IEEEmembership{Senior Member,~IEEE}
\thanks{Zhang Chen, Shuai Wan and Yuezhe Zhang are with the School of Electronics and Information, Northwestern Polytechnical University, Xi’an 710129, China (e-mail: chenzhang@mail.nwpu.edu.cn; swan@nwpu.edu.cn; yuezheZhang@mail.nwpu.edu.cn).}
\thanks{Siyu Ren and Junhui Hou are with the Department of Computer Science, City University of Hong Kong, Hong Kong SAR (e-mail: siyuren2-c@my.cityu.edu.hk; jh.hou@cityu.edu.hk).}
\thanks{Fuzheng Yang is with the School of Telecommunication Engineering, Xidian University, Xi’an 710071, China (e-mail: fzhyang@mail.xidian.edu.cn).}
\thanks{This work was supported in part by the TCL Science and Technology Innovation Fund, \ADDCZAQ{the NSFC under Grants 62371358 and 62422118, and the Hong Kong Research Grants Council under Grants 11219324 and N\_CityU1114/25.}}

}
% The paper headers
\markboth{}%
{Shell \MakeLowercase{}
}
\maketitle
\begin{abstract}
The unstructured and irregular nature of points poses a significant challenge for accurate point cloud quality assessment (PCQA), particularly in establishing accurate perceptual feature correspondence. To tackle this, we propose the Multi-scale Implicit Structural Similarity Measurement (MS-ISSM). Unlike traditional point-to-point matching, MS-ISSM utilizes radial basis function (RBF) to represent local features continuously, transforming distortion measurement into a comparison of implicit function coefficients. This approach effectively circumvents matching errors inherent in irregular data. Additionally, we propose a ResGrouped-MLP quality assessment network, which robustly maps multi-scale feature differences to perceptual scores. The network architecture departs from traditional flat multi-layer perceptron (MLP) by adopting a grouped encoding strategy integrated with residual blocks and channel-wise attention mechanisms. This hierarchical design allows the model to preserve the distinct physical semantics of luma, chroma, and geometry while adaptively focusing on the most salient distortion features across High, Medium, and Low scales. Experimental results on multiple benchmarks demonstrate that MS-ISSM outperforms state-of-the-art metrics in both reliability and generalization. The source code is available at: \textcolor{magenta}{https://github.com/ZhangChen2022/MS-ISSM}.
\end{abstract}

\begin{IEEEkeywords}
point cloud, quality assessment, multi-scale, implicit representation, multi-layer perceptron
\end{IEEEkeywords}
\section{Introduction} 
\IEEEPARstart{P}{oint} clouds are fundamental to 3D representation in applications ranging from autonomous driving to augmented and virtual reality (AR/VR) \cite{ref1,ref2,refsbAQ1,refsbAQ2,refsbAQ3}. However, their irregular and unstructured nature makes accurate quality assessment challenging, especially given distortions from noise, downsampling, and compression \cite{ref6}. While subjective assessment provides reliable ground truth, it is costly and time-consuming, necessitating efficient objective metrics that correlate well with human perception \cite{ref7,ref8,new1}.\par
\ADDCZ{Existing objective point cloud quality assessment (PCQA) methods generally fall into two categories:} projection-based and point-based\cite{ref11}. Projection-based methods project 3D data onto 2D planes, leveraging mature image quality assessment (IQA) algorithms \cite{ref12,ref14,ref15}. However, this dimensionality reduction often causes geometry loss and introduces viewpoint dependency \cite{ref16}. Conversely, point-based methods directly utilize 3D spatial features \cite{ref18,ref19}. A common approach involves identifying point-to-point correspondences via nearest-neighbor search to compute geometry or attribute distortions, 
%Yet, due to the unstructured nature of point clouds, establishing accurate correspondence is difficult, and discrete point errors often fail to reflect the continuous surface variations perceived by the human visual system (HVS) \cite{ref21,ref22}. \par
\ADDCZ{which introduces two main limitations. First, the unstructured nature of point clouds makes accurate correspondence difficult. Distortions like downsampling or \ADDCZ{noises often change the} density and spatial distribution \ADDCZ{of points}, leading to matching misalignment and calculation errors. Second, there is a discrepancy between discrete point-to-point errors and actual human visual perception. Traditional point-based metrics evaluate localized differences in isolation and fail to capture continuous structural changes \ADDCZ{on the surface}. Consequently, different distortion types can yield similar point-to-point error values but exhibit varying perceptual degradation to the human visual system (HVS)\cite{ref21,ref22}.} \par

\begin{figure}[t]
\centering  
\includegraphics[width=3.6in]{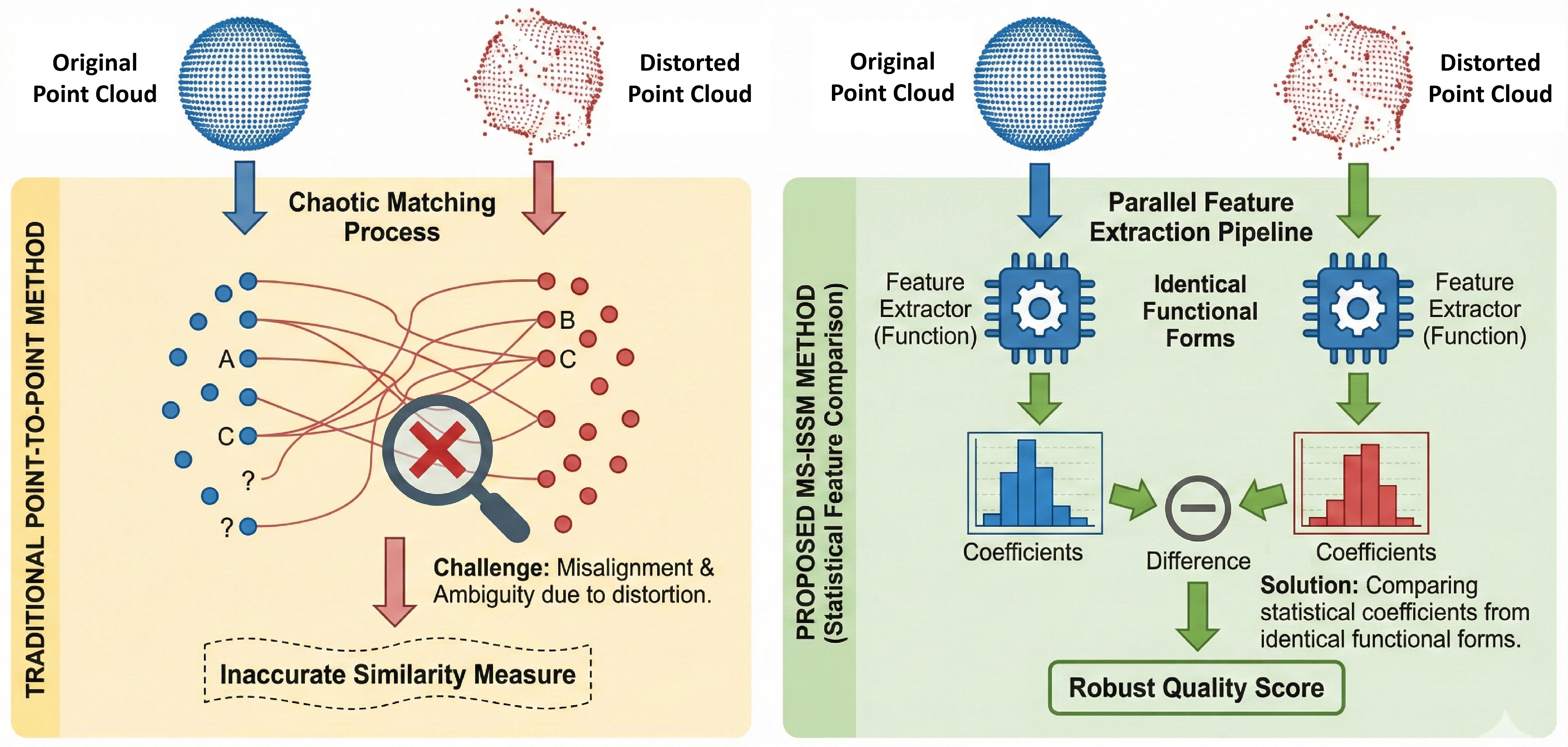}
\caption{\ADD{The difference between the MS-ISSM and the traditional point-to-point method.}}
\label{fig1}
\end{figure}
To address these limitations, we propose the multi-scale implicit structural similarity method (MS-ISSM). Building upon our previous work utilizing radial basis function (RBF) interpolation \cite{ref20}, \ADDCZ{we represent point cloud features as continuous implicit functions rather than discrete points.} 
\ADDCZ{As shown in Fig. \ref{fig1}, MS-ISSM models local spatial features as implicit functions and measures distortion by comparing their function coefficients. This approach bypasses the error-prone point matching process by evaluating localized structural topologies as a whole. Moreover, differences \ADDCZ{in coefficients} can reflect the continuous surface variations and structural degradation. Finally, to handle complex non-linear mappings, we integrate a multi-scale feature extraction strategy and propose a ResGrouped-MLP regression network to map these multi-scale coefficient differences to subjective quality scores. } 
The main contributions of this paper are summarized as follows: 
\begin{itemize} 
\item{We propose representing point cloud features using RBF implicit functions. By converting \ADDCZ{discrete point-based} feature differences into implicit function coefficient differences, we mitigate the accumulation of matching errors caused by the irregular nature of point clouds.} 
\item{\ADD{We design a ResGrouped-MLP quality assessment network. It incorporates a Log-Modulus transformation to handle heavy-tailed feature distributions and integrates residual blocks with channel-wise attention to adaptively weight physical semantics (luma, chroma, geometry) across multiple scales.}} 
\item{Extensive experiments on public datasets demonstrate that MS-ISSM achieves competitive performance \ADDCZ{in both reliability and generalization} compared to the state-of-the-art PCQA metrics.} \end{itemize} \par 

The remainder of this paper is organized as follows: Section \ref{sec2} reviews related work. Section \ref{sec3} details the problem formulation and theoretical foundation. Section \ref{sec4} describes the proposed MS-ISSM. Section \ref{sec6} presents experimental results, and Section \ref{sec7} concludes the paper.
\section{Related Work}\label{sec2} 
This section reviews existing PCQA methods, categorized into single-scale and multi-scale approaches.\par
{a) Single-scale PCQA Methods:} Single-scale methods quantify distortion by measuring geometric or attribute variations between corresponding points. Standard metrics rely on point-to-point Euclidean distances or feature differences, widely adopted in compression standards \cite{ref11}. Enhancements to these metrics include measuring projection distance along normal directions \cite{ref18}, utilizing Mahalanobis distance to capture spatial distribution \cite{ref25}, or calculating point-to-grid distances \cite{ref26}. To improve perceptual correlation, other approaches focus on feature disparities, such as angular differences between normal vectors \cite{ref27} or curvature variations \cite{ref28}. While these algorithms possess low computational complexity \cite{ref29}, they often fail to align with human visual perception due to the lack of perceptual modeling.\par
{b) Multi-scale PCQA Methods:} To better approximate human perception, researchers have integrated multi-scale and joint features. Meynet et al. proposed the Point Cloud Quality Metric (PCQM) \cite{ref30,ref19}, a linear combination of curvature, chroma, and brightness. Other hand-crafted feature methods combine geometric statistics with local plane features \cite{ref31}, utilize gradients from local graphs \cite{ref32,ref33}, analyze geometric topology alongside color distribution \cite{ref34}, or measure multi-scale spatial potential energy \cite{ref36}, transformational complexity \cite{ref37} and perception-guided hybrid metrics (PHM) \cite{refadd2}. Additionally, Lazzarotto et al. developed MS-PointSSIM by weighting structural similarity across spatial scales \cite{refadd1}.\par
Recent advancements leverage learning-based frameworks. These include CNN-based mapping of feature differences \cite{refadd3}, and GNNs for learning local intrinsic dependencies \cite{refadd4}. Other works employ PCA on local neighborhoods \cite{refadd5} or integrate spherical graph wavelet (SGW) coefficients with support vector regression (FRSVR) \cite{refadd7,refadd8}. Similarly, Cui et al. combined projected structural similarity with wavelet sub-band features in a learning framework \cite{refadd6}. Wang et al. also explored joint assessment using multi-scale texture features from 2D images and 3D points \cite{ref35}. \par
Alternatively, projection-based methods evaluate quality by rendering point clouds into 2D images and applying image quality assessment (IQA) models \cite{ref38,ref39,ref40}. However, projection alters 3D characteristics, leading to the loss of geometric details such as depth and occlusion relationships. Furthermore, while point-based learning methods show promise, they struggle with the fundamental challenge of establishing accurate point correspondences for distorted data.\par
\section{Problem Formulation}\label{sec3}
\subsection{Points Correspondence and Perceptual Distortion Measurement}
A point cloud is defined as a set of geometric coordinates and associated attributes. Let the original point cloud ${{\bf{P}}^{\rm{O}}}$ and the distorted point cloud ${{\bf{P}}^{\rm{D}}}$ be represented as:
\begin{equation}
\begin{array}{*{20}{c}}
{{{\bf{P}}^\alpha } = \left\{ {{\bf{p}}_n^\alpha ,{\bf{q}}_n^\alpha } \right\}_{n = 1}^{{N_\alpha }}}
\end{array}
\text{,}
\end{equation}
where $\alpha \in \{ {\rm{O}},{\rm{D}}\}$ denotes the type of point clouds, and $N_\alpha$ is the number of points. Each element consists of geometric coordinates ${\bf{p}}_n^\alpha$ and attributes ${\bf{q}}_n^\alpha$. Ideally, the perceptual distortion $D({{\bf{P}}^{\rm{O}}},{{\bf{P}}^{\rm{D}}})$ is measured by finding a feature bijection $\psi$ that minimizes the feature difference:
\begin{equation}
\setlength{\arraycolsep}{1.5pt} 
\renewcommand{\arraystretch}{1.2} 
\begin{array}{l}
D({{\bf{P}}^{\rm{O}}},{{\bf{P}}^{\rm{D}}})\\
= \mathop {\min }\limits_{\psi :{{\bf{P}}^{\rm{O}}} \to {{\bf{P}}^{\rm{D}}}} \left\{ {\frac{1}{{{N_\psi }}}\sum\limits_{{\bf{p}}_i^{\rm{O}} \in {{\bf{P}}^{\rm{O}}}}^{} {{{\left\| {{{\rm{M}}_{\rm{O}}}({\bf{p}}_i^{\rm{O}}) - {{\rm{M}}_{\rm{D}}}\left( {\psi \left( {{\bf{p}}_i^{\rm{O}}} \right)} \right)} \right\|}_2}} } \right\}
\end{array}
\text{,}\label{bijection}
\end{equation}
where ${{\rm{M}}_\alpha }(\cdot)$ extracts features (e.g., geometry, color). However, since ${N_{\rm{O}}}$ often differs from ${N_{\rm{D}}}$, a strict bijection is impractical. Consequently, classical methods approximate this using nearest-neighbor search to compute the symmetric distortion:
\begin{equation}
\setlength{\arraycolsep}{1.5pt} 
\renewcommand{\arraystretch}{1.2} 
\begin{array}{l}
\left\{ \begin{array}{l}
{D_{classic}}({{\bf{P}}^{\rm{O}}},{{\bf{P}}^{\rm{D}}}) = \max \left\{ {{d_{{\rm{O}} \to {\rm{D}}}},{d_{{\rm{D}} \to {\rm{O}}}}} \right\}\\
{d_{{\rm{O}} \to {\rm{D}}}} = \frac{1}{{{N_{\rm{O}}}}}\sum\limits_{i = 1}^{{N_{\rm{O}}}} {{{\left\| {{{\rm{M}}_{\rm{O}}}{\bf{(p}}_i^{\rm{O}}) - {{\rm{M}}_{\rm{D}}}\left( {{\varphi _{{\rm{O}} \to {\rm{D}}}}\left( {{\bf{p}}_i^{\rm{O}}} \right)} \right)} \right\|}_2}} \\
{d_{{\rm{D}} \to {\rm{O}}}} = \frac{1}{{{N_{\rm{D}}}}}\sum\limits_{j = 1}^{{N_{\rm{D}}}} {{{\left\| {{{\rm{M}}_{\rm{D}}}{\bf{(p}}_j^{\rm{D}}) - {{\rm{M}}_{\rm{O}}}\left( {{\varphi _{{\rm{D}} \to {\rm{O}}}}\left( {{\bf{p}}_j^{\rm{D}}} \right)} \right)} \right\|}_2}} 
\end{array} \right.
\end{array}
\text{,}\label{classic}
\end{equation}
where ${\varphi _{{\rm{O}} \to {\rm{D}}}}$ denotes the injective mapping determined by nearest-neighbor search. Due to the unordered nature of point cloud data and the random distribution of points in space, a simple nearest-neighbor search may result in incorrect mapping. For example, if the points in \ADDCZ{${{\bf{P}}^{\rm{O}}}$} are denser than those in \ADDCZ{${{\bf{P}}^{\rm{D}}}$} or if there is a spatial distribution bias between the two, the nearest-neighbor search leads to inaccurate matches. This, in turn, would affect the calculation of feature differences and, ultimately, the distortion measurement. To achieve accurate feature correspondence, in our earlier work\cite{ref20}, we obtained a bijective set of point features by using a feature interpolation function. However, this method only utilized single-scale luminance values. Additionally, \ADDCZ{the point-based methods} struggle to account for changes in the local structure of the point cloud, leading to distortion results that differ from actual perception.
\begin{figure*}[t]
\centering  
\includegraphics[width=5.8in]{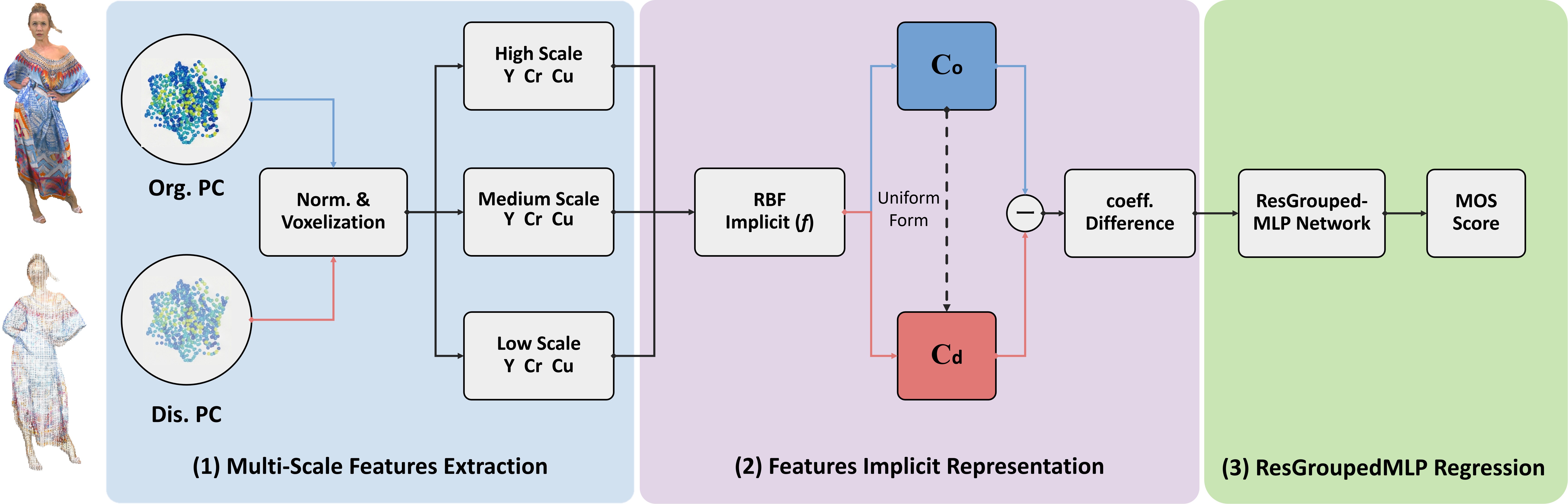}
\caption{\ADD{The schematic diagram depicts implementing the MS-ISSM solution. (1) Multi-scale features are extracted from the normalized distorted and original point clouds. The chroma, luma, and curvature features of each point cloud are calculated under high-, medium-, and low-quality conditions. (2) The RBF implicit representation is used to calculate the coefficient values for each feature, and multi-scale feature coefficient differences are calculated. (3) The ResGrouped-MLP is designed to map the multi-scale features coefficient differences to perceptual quality scores.}}
\label{overview}
\end{figure*}
\subsection{Feature Implicit Representation and Multi-scale Perceptual Distortion Calculation}
To address the above challenges, considering that exact point-to-point correspondence is difficult to achieve, for a local space, we map the feature set of the point cloud to a feature function, as shown in:
\begin{equation}
\setlength{\arraycolsep}{1.5pt} 
\renewcommand{\arraystretch}{1.2} 
\left\{ \begin{array}{l}
\left\{ {{\bf{p}}_i^{\rm{O}},{{\rm{M}}_{\rm{O}}}\left( {{\bf{p}}_i^{\rm{O}}} \right)} \right\}_{i = 1}^{{N_{\rm{O}}}} \to {f^{\rm{O}}}\left( {{\bf{p}}_i^{\rm{O}}{\rm{,}}{{\bf{W}}^{\rm{O}}}} \right) = {{\rm{M}}_{\rm{O}}}\left( {{\bf{p}}_i^{\rm{O}}} \right)\\
\left\{ {{\bf{p}}_j^{\rm{D}},{{\rm{M}}_{\rm{D}}}\left( {{\bf{p}}_j^{\rm{D}}} \right)} \right\}_{j = 1}^{{N_{\rm{D}}}} \to {f^{\rm{D}}}\left( {{\bf{p}}_j^{\rm{D}}{\rm{,}}{{\bf{W}}^{\rm{D}}}} \right) = {{\rm{M}}_{\rm{D}}}\left( {{\bf{p}}_j^{\rm{D}}} \right)
\end{array} \right.
\text{,}
\end{equation}
where ${f^{\rm{O}}}\left( {{\bf{p}}_i^{\rm{O}}{\rm{,}}{{\bf{W}}^{\rm{O}}}} \right)$ and ${f^{\rm{D}}}\left( {{\bf{p}}_j^{\rm{D}}{\rm{,}}{{\bf{W}}^{\rm{D}}}} \right)$ represent the feature functions of the original and distorted point clouds, respectively. ${{\bf{W}}^{\rm{O}}} = \left\{ {{w_k}^{\rm{O}}} \right\}_{k = 1}^K$ and ${{\bf{W}}^{\rm{D}}} = \left\{ {{w_k}^{\rm{D}}} \right\}_{k = 1}^K$ are the coefficient matrices of the implicit functions of ${{\bf{P}}^{\rm{O}}}$ and ${{\bf{P}}^{\rm{D}}}$, respectively. This approach transforms the feature difference into the error between their corresponding feature functions.\par
Since ${{\bf{W}}^{\rm{O}}} = \left\{ {{w_k}^{\rm{O}}} \right\}_{k = 1}^K$ and ${{\bf{W}}^{\rm{D}}} = \left\{ {{w_k}^{\rm{D}}} \right\}_{k = 1}^K$ have the same functional form, we calculate the feature function difference by the implicit function coefficients, as shown in:
\begin{equation}
\left\{ \begin{array}{l}
\ADDCZ{D'({{\bf{P}}^{\rm{O}}},{{\bf{P}}^{\rm{D}}}) = g\left( {{{d}_1},...,{{d}_k},...,{{d}_K}} \right)}\\
\ADDCZ{{d_k} = \left| {\frac{{w_k^{\rm{O}} - w_k^{\rm{D}}}}{{\max \left\{ {\left| {w_k^{\rm{O}}} \right|,\left| {w_k^{\rm{D}}} \right|} \right\} + \tau }}} \right|}
\end{array} \right.
\text{,}\label{eq5}
\end{equation}
where \ADDCZ{$\tau = 10^{-5}$ prevents division by zero,} and ${d_k}$ represents the difference between the individual function coefficients, and $g()$ is the nonlinear mapping from the feature function coefficient differences to the perceptual difference, and this mapping is obtained through a regression model. This method addresses the difficulty of point-to-point correspondence matching while considering the structural changes in local features.\par
Furthermore, considering the multi-scale nature of human visual perception, we incorporate the differences at low, medium, and high scales into the final distortion calculation. The final distortion is expressed as:
\begin{equation}
\renewcommand{\arraystretch}{1.2} 
{D_{pro}} = g\left( {\underbrace {{{d}_1},...,{{d}_K}}_{\rm{L}},\underbrace {{{d}_1},...,{{d}_K}}_{\rm{M}},\underbrace {{{d}_1},...,{{d}_K}}_{\rm{H}}} \right)
\text{.}\label{eq6}
\end{equation}
%%%%%%%%%%%%%%%%%%%%%%%%%%%%%%%%%%%%%%%%%%%%%%%%%%%%%%%%%%%%%%%%%%%%%%
\par

\begin{figure*}[h]
\centering  
\includegraphics[width=6.0in]{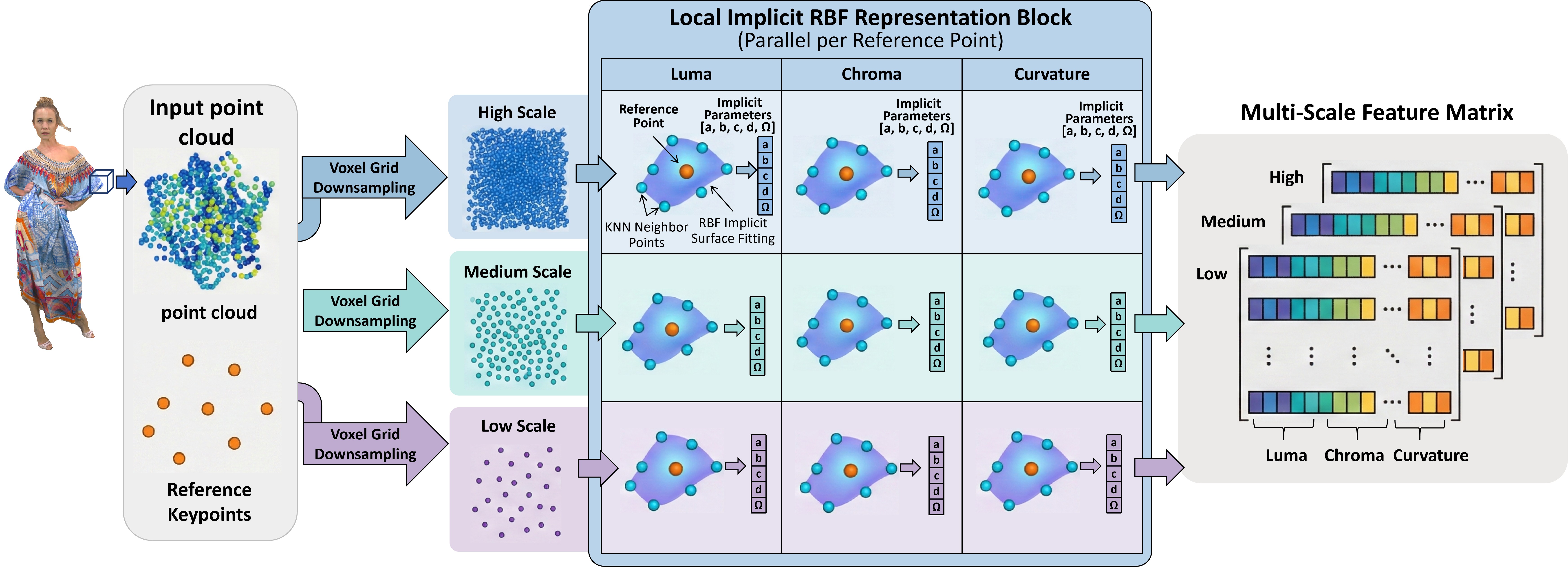}
\caption{\ADD{Schematic illustration of the Multi-scale implicit feature extraction framework for point cloud.}}
\label{multifeature}
\end{figure*}
\section{Proposed MS-ISSM}\label{sec4}
In this section, we first present the general framework of the proposed MS-ISSM in subsection $A$. Then, the implementation details of each module are described in subsections $B-D$, respectively.
\subsection{Overview} \label{sec4.1}
Due to the complexity of the human visual system (HVS), extracting features from point clouds and mapping them to precise perceptual quality metrics is a challenging task\cite{ref29}. To address this issue and streamline the computation process, we approach it in three steps.\par
In the first step, we extract multi-scale features from the normalized distorted and original point clouds. We focus on the chroma, luma, and curvature features of each point cloud under high-, medium-, and low-quality conditions. \par
In the second step, we apply the RBF implicit representation to the spatial scale features of the obtained distorted point cloud and the original point cloud, calculating the coefficient values for each feature's implicit representation.\par
\ADD{In the third step, we propose the ResGrouped-MLP to map the multi-scale feature coefficient differences to perceptual quality scores.}\par
The overall process of the proposed MS-ISSM is illustrated in Fig. \ref{overview}.\par
\subsection{Multi-scale Features Extraction}
To capture complex perceptual changes, we utilize three physical features that align with the HVS: curvature, luma, and chroma. Curvature describes the local surface geometry, reflecting sensitivity to both fine details and global structure \cite{ref28}. Luma and chroma, calculated from the point cloud's color components \cite{ref11}, represent light intensity and color distribution, respectively. \par
To ensure generalization across varying geometric scales, we normalize the geometric components of both distorted and original point clouds as follows:
\begin{equation}
{\bf{\hat p}}_{n}^{\alpha} = \frac{1024 \cdot ({\bf{p}}_{n}^{\alpha}-{\bf{p}}_{min})}{L_{max}}
\text{,}
\label{normalized}
\end{equation}

where ${\bf{\hat p}}_n^\alpha$ denotes the normalized coordinates. $L_{max}$ is the maximum edge length of the bounding box of the original point cloud ${{\bf{P}}^O}$, and ${{\bf{p}}_{\min }}$ is the coordinate-wise minimum vector. The resulting normalized point cloud is denoted as ${{\bf{\hat P}}^\alpha } = \left\{ {{\bf{\hat p}}_n^\alpha ,{\bf{q}}_n^\alpha } \right\}$.\par
Furthermore, inspired by the multi-layered perceptual mechanism of human vision, multi-scale point clouds are generated from both the original and distorted inputs through voxel grid downsampling. The voxel sizes are set to 2.0, 4.0, and 8.0, respectively. \ADDCZ{The base voxel size of 2.0 is selected as the trade-off between capturing high-frequency structural details and maintaining computational efficiency.} These values follow a dyadic progression, consistent with the hierarchical octree decomposition widely used in point cloud compression standards \cite{ref11}. This geometric progression allows for a systematic separation of spatial frequency components: the smallest scale captures high-frequency details (e.g., texture and noise), while the largest scale retains low-frequency structural information (e.g., global shape), ensuring a comprehensive evaluation of perceptual quality. Mimicking the way human vision adapts to different environments, this method focuses on various levels of detail, as shown in Fig. \ref{multifeature}.
%%%%%%%%%%%%%%%%%%%%%%%%%%%%%%%%%%%%%%%%%%%%%%%%%%%%%%%%%%%%%%%%%%%%%%%%%%
\begin{figure*}[h]
\centering  
\includegraphics[width=7.0in]{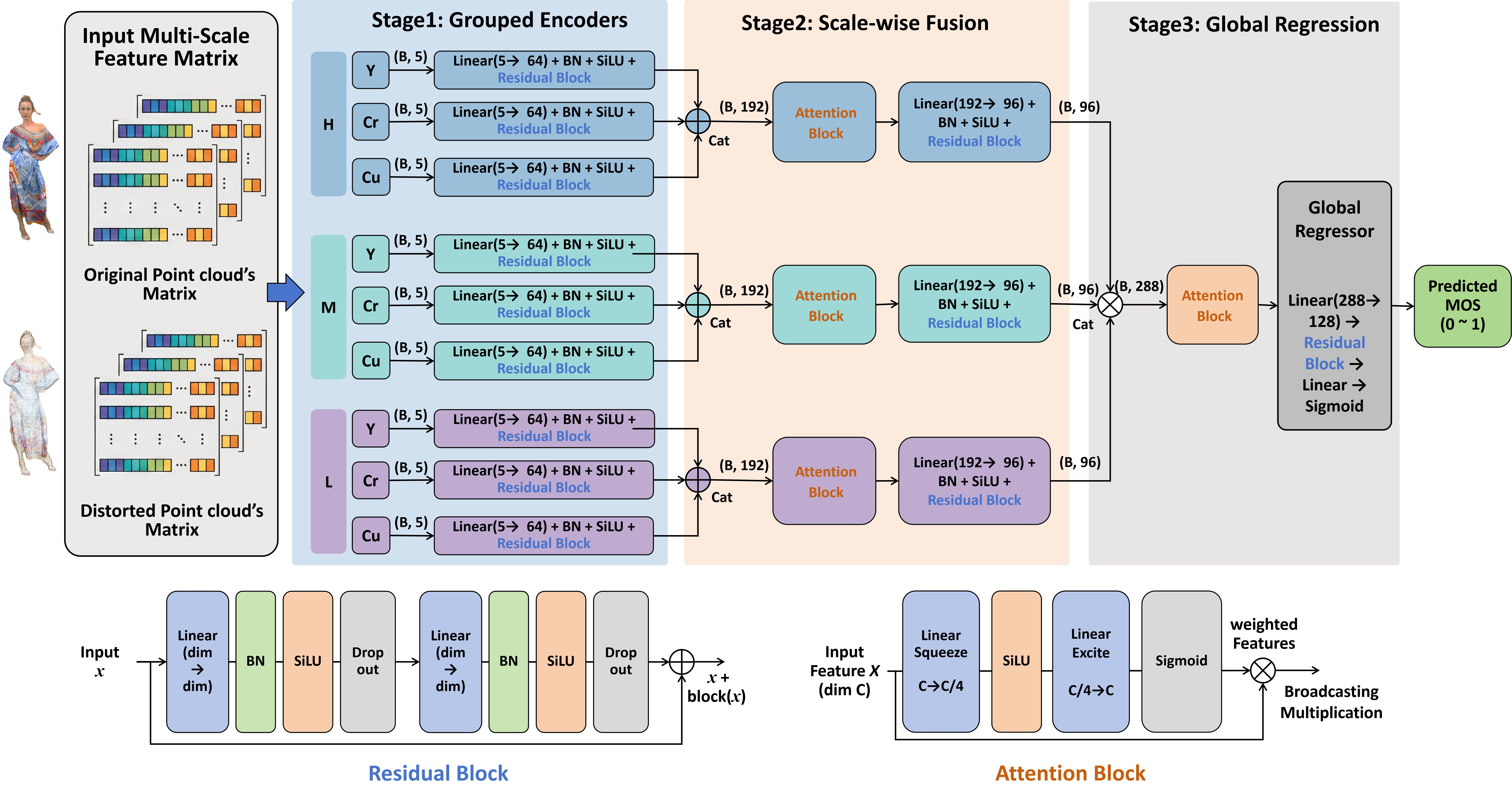}
\caption{\ADD{The Proposed ResGrouped-MLP Network.}}
\label{MLP network}
\end{figure*}

\subsection{Feature Implicit Representation}
The features of the local point cloud are implicitly represented using RBF \cite{ref41}:\par
\begin{equation}
\begin{array}{l}
f_{\bf{F}}^{\alpha ,\beta }({\bf{\hat p}}_{}^{\alpha ,\beta })\\
 = \eta _{\bf{F}}^{\alpha ,\beta }({\bf{\hat p}}_{}^{\alpha ,\beta }) + \sum\limits_{n = 1}^{N_\beta ^\alpha } {\left[ {\omega _{{\bf{F}},n}^{\alpha ,\beta } \cdot {\phi ^{\alpha ,\beta }}\left( {{{\left\| {{\bf{\hat p}}_{}^{\alpha ,\beta } - {\bf{\hat p}}_n^{\alpha ,\beta }} \right\|}_2}} \right)} \right]} 
\end{array}
\label{RBF} \text{,}
\end{equation}
where $\beta  \in \{ {\rm{H}},{\rm{M}},{\rm{L}}\}$ represents \ADDCZ{the type of} different spatial scales: high, medium, and low. ${\bf{F}} \in \{ \rm{Cu, Y, Cr}\} $ denotes feature types: curvature, luma, and chroma, and ${N^{\alpha,\beta }}$ indicates the number of points that influence the implicit function. $f_{\bf{F}}^{\alpha ,\beta }({\bf{\hat p}}_n^{\alpha ,\beta })$ corresponds to the implicit function associated with the current feature ${{\bf{F}}^{\alpha ,\beta }}$, and ${\bf{\hat p}}^{\alpha ,\beta } \in {{\bf{\hat p}}^{\alpha ,\beta }}$. $\omega _{{\bf{F}},n}^{\alpha ,\beta }$ denote the weight coefficients. 
\ADDCZ{${\phi ^{\alpha ,\beta }}\left( \right)$ denotes the RBF kernel function, defined as }
\begin{equation}
\ADDCZ{\phi^{\alpha,\beta} \left( \left\| \mathbf{\hat{p}}^{\alpha,\beta} - \mathbf{\hat{p}}_n^{\alpha,\beta} \right\|_2 \right) = \exp \left( -0.5 \left\| \mathbf{\hat{p}}^{\alpha,\beta} - \mathbf{\hat{p}}_n^{\alpha,\beta} \right\|_2^2 \right)
\text{.}}
\label{kernel}
\end{equation}
\ADDCZ{The Gaussian kernel function is chosen by its effectiveness and stability for localized point cloud feature representation\cite{ref20}.}
And $\eta _{\bf{F}}^{\alpha,\beta }({\bf{\hat p}}^{\alpha,\beta })$ is a three-variable polynomial with a maximum degree of 3. It is commonly expressed as 
\begin{equation}
\eta _{\bf{F}}^{\alpha ,\beta }({\bf{\hat p}}^{\alpha ,\beta }) = {\rm{a}}_{\bf{F}}^{\alpha ,\beta } \cdot \hat x^{\alpha ,\beta } + {\rm{b}}_{\bf{F}}^{\alpha ,\beta } \cdot \hat y^{\alpha ,\beta } + {\rm{c}}_{\bf{F}}^{\alpha ,\beta } \cdot \hat z^{\alpha ,\beta } + {\rm{d}}_{\bf{F}}^{\alpha ,\beta }
\text{,}
\end{equation}
where ${\rm{a}}_{\bf{F}}^{\alpha ,\beta }{\rm{,b}}_{\bf{F}}^{\alpha ,\beta }{\rm{,c}}_{\bf{F}}^{\alpha ,\beta }{\rm{,d}}_{\bf{F}}^{\alpha ,\beta }$ are constant coefficients. $\hat x_{}^{\alpha ,\beta }$, $\hat y_{}^{\alpha ,\beta }$, and $\hat z_{}^{\alpha ,\beta }$ represent coordinates of point ${{{\bf{\hat p}}}^{\alpha ,\beta }}$ at $x, y, z$ directions, respectively. \par
\ADDCZ{To ensure a non-singular coefficient matrix for a unique solution and prevent the RBF kernel from interfering with the polynomial terms,} the weight coefficients $\omega _{{\bf{F}},n}^{\alpha ,\beta }$ must satisfy the following constraint conditions:
\begin{equation}
\resizebox{1.0\hsize}{!}{$
\sum\limits_{n = 1}^{{N^{\alpha ,\beta }}} {\omega _{{\bf{F}},n}^{\alpha ,\beta }} = \sum\limits_{n = 1}^{{N^{\alpha ,\beta }}} {\omega _{{\bf{F}},n}^{\alpha ,\beta }\hat x_n^{\alpha ,\beta }} = \sum\limits_{n = 1}^{{N^{\alpha ,\beta }}} {\omega _{{\bf{F}},n}^{\alpha ,\beta }\hat y_n^{\alpha ,\beta }} = \sum\limits_{n = 1}^{{N^{\alpha ,\beta }}} {\omega _{{\bf{F}},n}^{\alpha ,\beta }\hat z_n^{\alpha ,\beta }} = 0
$}
\label{constraint}\text{.}
\end{equation}
And by inputting all coordinates of ${{{\bf{\hat p}}}^{\alpha ,\beta }}$ in ${{\bf{\hat P}}^{\alpha ,\beta }}$ into Eq. (\ref{RBF}), we determine the coefficients of $\eta _{\bf{F}}^{\alpha ,\beta }({\bf{\hat p}}_{}^{\alpha ,\beta })$ and $\omega _{{\bf{F}},n}^{\alpha ,\beta }$ through the following equation:  
 \begin{equation}
{\mathbf{X}^{\alpha ,\beta }} \cdot \mathbf{W}_{\bf{F}}^{\alpha ,\beta } = {\mathbf{Y}_{\bf{F}}^{\alpha ,\beta }}
\label{matrix} \text{.}
\end{equation} 
\ADDCZ{In our implementation, we use Householder QR decomposition to solve Eq. (\ref{matrix}), preventing the numerical instability of direct matrix inversion.} $\mathbf{Y}_{\bf{F}}^{\alpha ,\beta }$ is the feature matrix, as shown in 
\begin{equation}
\mathbf{Y}_{\bf{F}}^{\alpha ,\beta } = {\left[ {\begin{array}{*{20}{c}}
{{{\bf{F}}^{\alpha ,\beta }}({\bf{\hat p}}_1^{\alpha ,\beta })}& \cdots &{{{\bf{F}}^{\alpha, \beta }} ({\bf{\hat p}}_{{N^{\alpha ,\beta }}}^{\alpha ,\beta })}&{\mathbf{0}}
\end{array}} \right]^ \top }
\label{matrixY}\text{.}
\end{equation}
${{\mathbf{X}}^{\alpha ,\beta }}$ is the coordinate matrix, as represented in 
\begin{equation}
\setlength{\arraycolsep}{1.5pt} 
\renewcommand{\arraystretch}{1.2} 
\begin{array}{l}
{{\mathbf{X}}^{\alpha ,\beta}}
 = \left[ {\begin{array}{*{20}{c}}
{\phi _{11}^{\alpha ,\beta }}& \cdots &{\phi _{1{N^{\alpha ,\beta }}}^{\alpha ,\beta }}&{\hat x_1^{\alpha ,\beta }}&{\hat y_1^{\alpha ,\beta }}&{\hat z_1^{\alpha ,\beta }}&1\\
 \vdots & \ddots & \vdots & \vdots & \vdots & \vdots & \vdots \\
{\phi _{{N^{\alpha ,\beta }}1}^{\alpha ,\beta }}& \cdots &{\phi _{{N^{\alpha ,\beta }}{N^{\alpha ,\beta }}}^{\alpha ,\beta }}&{\hat x_{{N^{\alpha ,\beta }}}^{\alpha ,\beta }}&{\hat y_{{N^{\alpha ,\beta }}}^{\alpha ,\beta }}&{\hat z_{{N^{\alpha ,\beta }}}^{\alpha ,\beta }}&1\\
1& \cdots &1&0&0&0&0\\
{\hat x_1^{\alpha ,\beta }}& \cdots &{\hat x_{{N^{\alpha ,\beta }}}^{\alpha ,\beta }}&0&0&0&0\\
{\hat y_1^{\alpha ,\beta }}& \cdots &{\hat y_{{N^{\alpha ,\beta }}}^{\alpha ,\beta }}&0&0&0&0\\
{\hat z_1^{\alpha ,\beta }}& \cdots &{\hat z_{{N^{\alpha ,\beta }}}^{\alpha ,\beta }}&0&0&0&0
\end{array}} \right]
\end{array}
\label{matrixX}
\end{equation}
where ${\phi_{12}^{\alpha ,\beta }}$ is equal to ${\phi^{\alpha ,\beta } \left( {{{\left\| {{{\bf{\hat p}}_1^{\alpha ,\beta }} - {{\bf{\hat p}}_2^{\alpha ,\beta }}} \right\|}_2}} \right)}$. And $\mathbf{W}_{\bf{F}}^{\alpha ,\beta }$ is the weight matrix, as represented in 
\begin{equation}
\setlength{\arraycolsep}{1.5pt} 
\begin{array}{l}
\mathbf{W}_{\bf{F}}^{\alpha ,\beta }
 = {\left[ {\begin{array}{*{20}{c}}
{\omega _{{\bf{F}},1}^{\alpha ,\beta }}& \cdots &{\omega _{{\bf{F}},{N^{\alpha ,\beta }}}^{\alpha ,\beta }}&{{\rm{a}}_{\bf{F}}^{\alpha ,\beta }}&{{\rm{b}}_{\bf{F}}^{\alpha ,\beta }}&{{\rm{c}}_{\bf{F}}^{\alpha ,\beta }}&{{\rm{d}}_{\bf{F}}^{\alpha ,\beta }}
\end{array}} \right]^ \top }
\end{array}
\label{matrixW}\text{.}
\end{equation}  

Since the number of coefficients in the weight matrix ${\mathbf{W}}_{\bf{F}}^{\alpha,\beta }$ is determined by the number of points ${N^{\alpha,\beta }}$, to simplify the computational process and facilitate the comparison of distortions using the coefficients, we downsample $\bf{\hat P}^O$ to obtain a set of reference points, denoted as ${{\bf{P}}^{\rm{R}}} = \left\{ {{\bf{p}}_t^{\rm{R}}} \right\}_{t = 1}^{{N_{\rm{R}}}}$. Using nearest-neighbor search, for each point ${\bf{p}}_t^{\rm{R}}$ in the reference point set, we find the 30 closest neighbors in both $\left\{ {{\bf{\hat p}}_i^{\rm{O}}} \right\}_{i = 1}^{{N_{\rm{O}}}}$ and $\left\{ {{\bf{\hat p}}_j^{\rm{D}}} \right\}_{j = 1}^{{N_{\rm{D}}}}$, which are then used to compute the implicit function for the region around each reference point.
Finally, the features of ${\bf{\hat P}}^{\rm{O}}$ and ${\bf{\hat P}}^{\rm{D}}$ can be represented as tensors ${\mathbf{C} ^\alpha }$:
\begin{equation}
{\mathbf{C} ^\alpha } = \left[{\begin{array}{*{20}{c}}
{{\mathbf{W}}_{\rm{Cu}}^{\alpha ,\rm{H}}}&{{\mathbf{W}}_{\rm{Y}}^{\alpha ,\rm{H}}}&{{\mathbf{W}}_{\rm{Cr}}^{\alpha ,\rm{H}}}\\
{{\mathbf{W}}_{\rm{Cu}}^{\alpha ,\rm{M}}}&{{\mathbf{W}}_{\rm{Y}}^{\alpha ,\rm{M}}}&{{\mathbf{W}}_{\rm{Cr}}^{\alpha ,\rm{M}}}\\
{{\mathbf{W}}_{\rm{Cu}}^{\alpha ,\rm{L}}}&{{\mathbf{W}}_{\rm{Y}}^{\alpha ,\rm{L}}}&{{\mathbf{W}}_{\rm{Cr}}^{\alpha ,\rm{L}}}
\end{array}} \right]
\text{.}
\end{equation} 
By substituting $\mathbf{C}^{\rm{O}}$ and $\mathbf{C}^{\rm{D}}$ into Eq. (\ref{eq5}) and (\ref{eq6}), the quality score of ${\bf{P}}^{\rm{D}}$ is calculated. Considering the simplicity of the algorithm and to avoid an excessive number of comparison coefficients, we take the average of weight coefficients differences. And the function $g()$ is obtained through the ResGrouped-MLP network as follow.\par

\subsection{\ADD{ResGrouped-MLP Regression}}
\ADD{We propose the ResGrouped-MLP, a hierarchical deep learning framework designed to robustly map hand-crafted point cloud features to subjective quality scores. Addressing the limitations of flat networks, we adopt a ``Split-Transform-Merge" architecture to preserve the distinct physical semantics of features. The framework integrates a novel Log-Modulus preprocessing and a multi-scale attention mechanism, as shown in Fig. \ref{MLP network}.}\par
\begin{figure}[h]
\centering
\includegraphics[width=3.2in]{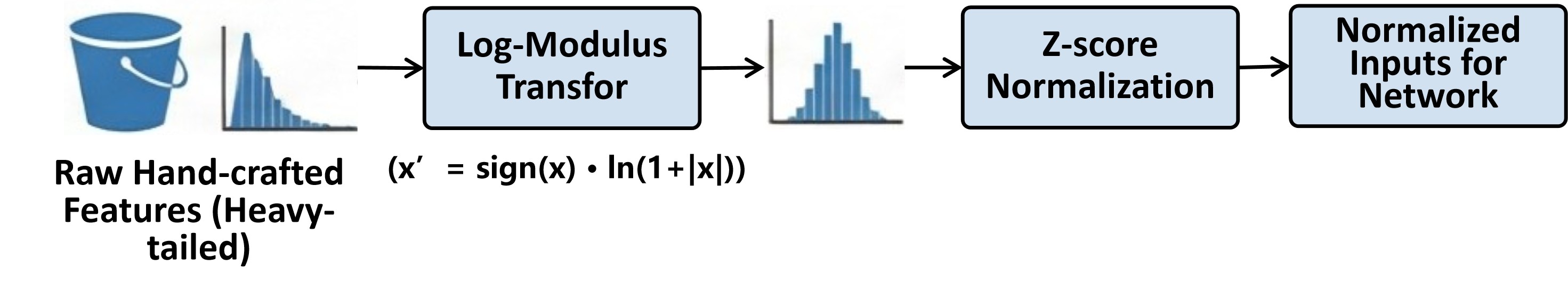}
\caption{\ADDCZ{Preprocessing pipeline of hand-crafted features}}
\label{preprocessing}
\end{figure}
{\bf{\ADD{Log-Modulus Preprocessing:}}} \ADD{Statistical features extracted from point clouds often exhibit a heavy-tailed distribution, where standard Z-score normalization can lead to gradient instability and poor convergence. To address this, \ADDCZ{we selected the Log-Modulus transformation to handle signed data and compress heavy-tailed distributions caused by local distortions. This parameter-free operation improves network training stability as follows:}
\begin{equation}
{\rm{x}'} = \text{sign}(\rm{x}) \cdot \ln(1 + |\rm{x}|)
\text{,}
\end{equation}
where $\rm{x}$ denotes the raw feature value. This transformation effectively rectifies the data distribution towards a quasi-normal form, allowing the subsequent network to focus on underlying feature patterns rather than being biased by outliers, as shown in Fig. \ref{preprocessing}.}\par
{\bf{\ADD{The ResGrouped-MLP Architecture:}}} \ADD{Point cloud quality perception relies on the interplay between feature scales (High, Medium, Low) and attribute channels (luma, chroma, curvature). Simply concatenating these features risks losing their distinct physical meanings. Therefore, we design a hierarchical architecture composed of three key stages:}\par
\ADD{a) {Deep Grouped Encoders:} Instead of a generic fully connected layer, we treat each scale-channel pair as an independent group. We employ Residual Blocks for feature encoding, formulated as $\rm{x} + \mathcal{F}(\rm{x})$, where $\mathcal{F}()$ represents a dual-layer perceptron with Batch Normalization and SiLU activation. This grouped design isolates the specific distortion characteristics of each channel, preventing information interference at early stages while mitigating the vanishing gradient problem.}\par
\ADD{b) {Scale-wise Attention Fusion:} Since different color channels contribute unequally to human perception, we introduce a Channel Attention Block to recalibrate features within each scale. The mechanism adaptively learns weights to highlight salient distortions:
\begin{equation}
\ADDCZ{\mathbf{f}{scale}' = \mathbf{f}{scale} \otimes \sigma(\text{MLP}(\mathbf{f}{scale}))}
\text{,}
\end{equation}
where $\mathbf{f}{scale}$ is the concatenated feature vector, $\otimes$ denotes element-wise multiplication, and $\sigma(\cdot)$ is the Sigmoid function. The internal MLP utilizes a bottleneck structure with a reduction ratio $r=4$. This bottleneck compresses the feature space to aggregate global information before restoring it, enabling the network to dynamically suppress noise and emphasize the most relevant feature channels.}\par
\ADD{c) {Global Hierarchical Regression:} Finally, the refined features from H, M, and L scales are concatenated and passed through a Global Attention module. This ensures the model captures the complex interaction between global geometry (Low scale) and local fine-grained details (High scale) before mapping them to the final quality score.}\par
{\bf{\ADD{Loss Function:}}} \ADD{To ensure prediction accuracy, linearity, and monotonic consistency, we utilize a hybrid loss function combining mean squared error (MSE), Pearson Linear Correlation Coefficient (PLCC) loss, and Margin Ranking loss:
\begin{equation}
\mathcal{L}{total} = \mathcal{L}{MSE} + \lambda_1 \mathcal{L}{PLCC} + \lambda_2 \mathcal{L}{Rank}
\text{,}
\end{equation}
where $\lambda_1$ and $\lambda_2$ are weighting parameters used to balance the optimization objectives.} \ADDCZ{And, these values are set to 0.2 and 0.2, respectively. Such a configuration ensures that the PLCC loss provides strong linear correlation guidance and the Margin Ranking loss penalizes monotonicity violations without overwhelming the primary MSE regression.}\par
{\bf{Implementation and Validation:}} The model is trained using the AdamW optimizer with a weight decay of $10^{-2}$ and a Cosine Annealing scheduler for 80 epochs (batch size 32). \ADDCZ{The detailed hyperparameter configurations, including the grouped encoders, scale-wise attention and global regressor, are summarized in Table \ref{tab:config}}.

\begin{table}[htbp]
\centering
\caption{\ADDCZ{Detailed Configurations of the ResGrouped-MLP.}}
\label{tab:network_config}
\renewcommand{\arraystretch}{1.3} 
\begin{tabularx}{\linewidth}{l >{\raggedright\arraybackslash}X}
\toprule
{Module} & {Sequential Operations} \\ 
\midrule
grouped encoders & FC (64), BN, SiLU, ResBlock (dropout = 0.1) \\ 
scale-wise attention & FC (48), SiLU, FC (192), sigmoid \\ 
global regressor & FC (128), BN, SiLU, ResBlock (dropout = 0.2), FC (64), SiLU, FC (1), sigmoid \\ 
\bottomrule
\label{tab:config}
\end{tabularx}
\end{table}

\section{Experimental Evaluations}\label{sec6}
This section uses four publicly available point cloud subjective datasets to validate the proposed method's effectiveness in perceptual evaluation. We compare the quality assessment results of our method with those of classic and state-of-the-art (SOTA) PCQA metrics. By analyzing the results across various datasets, we can assess the robustness and generalization of our method in different distortions. 

\begin{table}[htbp]
\centering
\setlength{\tabcolsep}{3pt} 
\renewcommand{\arraystretch}{0.85} 
\definecolor{bestred}{HTML}{FF0000}
\definecolor{secondblue}{HTML}{3531FF}
\caption{\ADDCZ{Performance comparison of different PCQA metrics. The best, second-best, and third-best among the primary methods are highlighted in \textcolor{bestred}{red}, \textcolor{secondblue}{blue}, and \underline{underlined}, respectively.}}
\begin{adjustbox}{max width=\linewidth}
\begin{tabular}{llccccc}
\toprule
\multirow{2}{*}{Method} & \multirow{2}{*}{Crit.} & \multicolumn{4}{c}{Datasets} & \multirow{2}{*}{Rank} \\ \cmidrule(l){3-6} 
 &  & \shortstack{SJTU\\\cite{ref14}} & \shortstack{ICIP\\\cite{ref43}} & \shortstack{M-PCCD\\\cite{ref42}} & \shortstack{WPC\\\cite{ref39}} & \\ \midrule
\multirow{4}{*}{\shortstack[l]{PSNR-p2p\\\cite{ref11}}} 
 & PLCC & 0.460 & 0.862 & 0.395 & 0.260 & 12.25 \\
 & SROCC & 0.398 & 0.853 & 0.239 & 0.008 & 12.00 \\
 & KROCC & 0.298 & 0.590 & 0.196 & 0.002 & 12.75 \\
 & RMSE & 2.130 & 0.659 & 1.250 & 22.137 & 12.50 \\ \midrule
\multirow{4}{*}{\shortstack[l]{PSNR-p2pl\\\cite{ref11}}} 
 & PLCC & 0.433 & 0.898 & 0.395 & 0.260 & 11.75 \\
 & SROCC & 0.415 & 0.843 & 0.232 & 0.053 & 12.00 \\
 & KROCC & 0.303 & 0.606 & 0.201 & 0.034 & 11.75 \\
 & RMSE & 2.162 & 0.642 & 1.250 & 22.137 & 11.75 \\ \midrule
\multirow{4}{*}{\shortstack[l]{PSNR-YUV\\\cite{ref11}}} 
 & PLCC & 0.550 & 0.779 & 0.603 & 0.387 & 11.50 \\
 & SROCC & 0.555 & 0.777 & 0.620 & 0.384 & 11.50 \\
 & KROCC & 0.385 & 0.580 & 0.442 & 0.263 & 11.50 \\
 & RMSE & 2.003 & 0.712 & 1.085 & 21.139 & 11.50 \\ \midrule
 \multirow{4}{*}{\shortstack[l]{PSSIM\\\cite{ref34}}} 
 & PLCC & 0.769 & 0.886 & 0.712 & 0.500 & 10.00 \\
 & SROCC & 0.717 & 0.877 & 0.716 & 0.424 & 9.75 \\
 & KROCC & 0.522 & 0.701 & 0.538 & 0.300 & 9.75 \\
 & RMSE & 1.533 & 0.653 & 0.956 & 19.854 & 10.25 \\ \midrule
 \multirow{4}{*}{\shortstack[l]{PCQM\\\cite{ref19}}} 
 & PLCC & 0.807 & \underline{0.948} & 0.903 & 0.563 & 6.75 \\
 & SROCC & 0.774 & \textcolor{secondblue}{0.956} & 0.916 & 0.545 & 6.50 \\
 & KROCC & 0.590 & 0.830 & 0.748 & 0.432 & 7.25 \\
 & RMSE & 1.418 & \underline{0.363} & 0.584 & 18.947 & 7.00 \\ \midrule
 \multirow{4}{*}{\shortstack[l]{MS-PSSIM\\\cite{refadd1}}} 
 & PLCC & 0.844 & 0.929 & 0.768 & 0.551 & 7.50 \\
 & SROCC & 0.842 & 0.906 & 0.777 & 0.541 & 7.50 \\
 & KROCC & 0.645 & 0.771 & 0.602 & 0.381 & 7.25 \\
 & RMSE & 1.286 & 0.612 & 0.872 & 19.132 & 8.25 \\ \midrule
  \multirow{4}{*}{\shortstack[l]{GraphSIM\\\cite{ref32}}} 
 & PLCC & 0.860 & 0.901 & 0.921 & 0.690 & 6.00 \\
 & SROCC & 0.843 & 0.888 & 0.936 & 0.679 & 6.25 \\
 & KROCC & 0.641 & 0.723 & 0.786 & 0.493 & 6.50 \\
 & RMSE & 1.223 & 0.492 & 0.529 & 16.597 & 6.00 \\ \midrule
\multirow{4}{*}{\shortstack[l]{MS-GraphSIM\\\cite{ref33}}} 
 & PLCC & 0.897 & 0.902 & 0.906 & 0.702 & 5.75 \\
 & SROCC & 0.874 & 0.890 & 0.922 & 0.701 & 5.75 \\
 & KROCC & 0.682 & 0.727 & 0.761 & 0.517 & 6.00 \\
 & RMSE & \underline{1.058} & 0.491 & 0.575 & 16.324 & 5.50 \\ \midrule
\multirow{4}{*}{\shortstack[l]{PHM\\\cite{refadd2}}} 
 & PLCC & \underline{0.901} & 0.867 & \textcolor{secondblue}{0.945} & \textcolor{secondblue}{0.838} & \underline{4.50} \\
 & SROCC & \underline{0.880} & 0.831 & \textcolor{secondblue}{0.948} & \textcolor{secondblue}{0.832} & 4.75 \\
 & KROCC & \underline{0.695} & 0.701 & \underline{0.807} & \textcolor{secondblue}{0.639} & 4.50 \\
 & RMSE & \textcolor{secondblue}{1.038} & 0.566 & 0.445 & \textcolor{secondblue}{12.509} & \underline{4.00} \\ \midrule
 \multirow{4}{*}{\shortstack[l]{TCDM\\\cite{ref37}}} 
 & PLCC & \textcolor{bestred}{0.931} & 0.942 & \underline{0.937} & 0.805 & \textcolor{secondblue}{3.50} \\
 & SROCC & \textcolor{bestred}{0.912} & 0.935 & 0.945 & 0.803 & \textcolor{secondblue}{3.75} \\
 & KROCC & \textcolor{bestred}{0.740} & 0.803 & 0.801 & 0.604 & \textcolor{secondblue}{3.75} \\
 & RMSE & \textcolor{bestred}{0.875} & 0.381 & 0.475 & 13.607 & \underline{4.00} \\  \midrule
\multirow{4}{*}{\shortstack[l]{FRSVR\\(K-fold splicing)\\\cite{refadd8}}} 
 & PLCC & 0.822 & \textcolor{secondblue}{0.957} & 0.886 & \underline{0.825} & 5.00 \\
 & SROCC & 0.806 & \underline{0.954} & 0.904 & \underline{0.827} & 5.25 \\
 & KROCC & 0.601 & \textcolor{secondblue}{0.843} & 0.783 & \underline{0.620} & 4.50 \\
 & RMSE & 1.333 & \textcolor{secondblue}{0.341} & \textcolor{bestred}{0.341} & \underline{12.786} & \textcolor{secondblue}{3.25} \\ \midrule
\multirow{4}{*}{\shortstack[l]{PointPCA\\(K-fold splicing)\\\cite{refadd5}}} 
 & PLCC & 0.819 & 0.942 & 0.912 & 0.806 & 5.25 \\
 & SROCC & 0.786 & 0.946 & \underline{0.946} & 0.807 & \underline{4.75} \\
 & KROCC & 0.592 & \underline{0.837} & \textcolor{secondblue}{0.827} & 0.607 & \underline{4.25} \\
 & RMSE & 1.341 & 0.375 & \underline{0.375} & 13.456 & 4.75 \\  \midrule
\multirow{4}{*}{\shortstack[l]{\textbf{MS-ISSM}\\\textbf{(K-fold splicing)}}} 
 & PLCC & \textcolor{secondblue}{0.906} & \textcolor{bestred}{0.964} & \textcolor{bestred}{0.958} & \textcolor{bestred}{0.855} & \textcolor{bestred}{1.25} \\
 & SROCC & \textcolor{secondblue}{0.897} & \textcolor{bestred}{0.967} & \textcolor{bestred}{0.961} & \textcolor{bestred}{0.846} & \textcolor{bestred}{1.25} \\
 & KROCC & \textcolor{secondblue}{0.698} & \textcolor{bestred}{0.852} & \textcolor{bestred}{0.853} & \textcolor{bestred}{0.666} & \textcolor{bestred}{1.25} \\
 & RMSE & 1.224 & \textcolor{bestred}{0.283} & \textcolor{secondblue}{0.351} & \textcolor{bestred}{11.736} & \textcolor{bestred}{2.25} \\ 
 \midrule
 \\
 \midrule
\multirow{4}{*}{\shortstack[l]{FRSVR\\(K-fold average)\\\cite{refadd8}}} 
 & PLCC & 0.858 & 0.976 & 0.916 & 0.842 & - \\
 & SROCC & 0.823 & 0.969 & 0.929 & 0.836 & - \\
 & KROCC & 0.622 & 0.893 & 0.778 & 0.647 & - \\
 & RMSE & 1.279 & 0.279 & 0.432 & 12.143 & - \\
 \midrule
\multirow{4}{*}{\shortstack[l]{PointPCA\\(K-fold average)\\\cite{refadd5}}} 
 & PLCC & 0.847 & 0.976 & 0.956 & 0.812 & - \\
 & SROCC & 0.842 & 0.974 & 0.964 & 0.815 & - \\
 & KROCC & 0.625 & 0.855 & 0.855 & 0.614 & - \\
 & RMSE & 1.293 & 0.269 & 0.339 & 13.078 & - \\ 
 \midrule
\multirow{4}{*}{\shortstack[l]{\textbf{MS-ISSM}\\\textbf{(K-fold average)}}} 
 & PLCC & 0.925 & 0.968 & 0.966 & 0.867 & - \\
 & SROCC & 0.911 & 0.976 & 0.971 & 0.859 & - \\
 & KROCC & 0.763 & 0.917 & 0.883 & 0.661 & - \\
 & RMSE & 1.043 & 0.256 & 0.344 & 11.618 & - \\
 \bottomrule
\end{tabular}
\end{adjustbox}
\label{table_all}
\end{table}

\begin{table}[htbp]
\centering
\setlength{\tabcolsep}{3pt} % 紧凑的列间距
\renewcommand{\arraystretch}{0.85} % 减小行高以适应单栏高度
\definecolor{bestred}{HTML}{FF0000}
\definecolor{secondblue}{HTML}{3531FF}
\caption{\ADDCZ{Performance comparison of classic and SOTA metrics on different distortion types. The best, second-best, and third-best are highlighted in \textcolor{bestred}{red}, \textcolor{secondblue}{blue}, and \underline{underlined}, respectively.}}
\begin{adjustbox}{max width=\linewidth}
\begin{tabular}{llcccccc}
\toprule
\multirow{2}{*}{Method} & \multirow{2}{*}{Crit.} & \multicolumn{6}{c}{Distortion types} \\ \cmidrule(l){3-8} 
 & & Ds & Ns & Oc & Mx & Tc & Vc \\ \midrule
\multirow{4}{*}{PSNR-p2p} 
 & PLCC  & 0.394 & 0.649 & 0.525 & 0.737 & 0.468 & 0.208 \\
 & SROCC & 0.333 & 0.649 & 0.368 & 0.673 & 0.440 & 0.196 \\
 & KROCC & 0.232 & 0.482 & 0.267 & 0.513 & 0.311 & 0.133 \\
 & RMSE  & 1.330 & 0.806 & 1.082 & 0.851 & 1.115 & 0.893 \\ \midrule
\multirow{4}{*}{PSNR-p2pl} 
 & PLCC  & 0.296 & 0.672 & 0.527 & 0.666 & 0.471 & 0.264 \\
 & SROCC & 0.188 & 0.668 & 0.373 & 0.636 & 0.437 & 0.243 \\
 & KROCC & 0.135 & 0.494 & 0.270 & 0.473 & 0.310 & 0.166 \\
 & RMSE  & 1.383 & 0.784 & 1.080 & 0.939 & 1.114 & 0.881 \\ \midrule
\multirow{4}{*}{PSNR-YUV} 
 & PLCC  & 0.632 & 0.588 & 0.561 & 0.670 & 0.541 & 0.393 \\
 & SROCC & 0.617 & 0.559 & 0.543 & 0.659 & 0.528 & 0.391 \\
 & KROCC & 0.421 & 0.404 & 0.364 & 0.490 & 0.369 & 0.267 \\
 & RMSE  & 1.122 & 0.856 & 1.052 & 0.934 & 1.062 & 0.840 \\ \midrule
\multirow{4}{*}{PSSIM} 
 & PLCC  & \textcolor{bestred}{0.959} & 0.771 & 0.728 & 0.881 & 0.464 & 0.303 \\
 & SROCC & \textcolor{bestred}{0.906} & 0.766 & 0.564 & 0.877 & 0.091 & 0.166 \\
 & KROCC & \textcolor{bestred}{0.719} & 0.585 & 0.431 & 0.691 & 0.062 & 0.112 \\
 & RMSE  & \textcolor{bestred}{0.413} & 0.675 & 0.872 & 0.595 & 1.118 & 0.870 \\ \midrule
\multirow{4}{*}{PCQM} 
 & PLCC  & 0.435 & \underline{0.900} & 0.870 & 0.902 & 0.718 & 0.643 \\
 & SROCC & 0.054 & \underline{0.902} & 0.872 & 0.881 & 0.714 & 0.623 \\
 & KROCC & 0.047 & \underline{0.720} & 0.683 & 0.691 & 0.519 & 0.443 \\
 & RMSE  & 1.303 & \underline{0.462} & 0.627 & 0.544 & 0.879 & 0.700 \\ \midrule
\multirow{4}{*}{MS-PSSIM} 
 & PLCC  & 0.703 & 0.758 & 0.753 & 0.880 & 0.611 & 0.359 \\
 & SROCC & 0.689 & 0.737 & 0.745 & 0.865 & 0.565 & 0.338 \\
 & KROCC & 0.516 & 0.539 & 0.551 & 0.675 & 0.399 & 0.233 \\
 & RMSE  & 1.029 & 0.690 & 0.836 & 0.597 & 0.999 & 0.852 \\ \midrule
\multirow{4}{*}{GraphSIM} 
 & PLCC  & 0.943 & 0.885 & 0.854 & 0.895 & 0.628 & 0.637 \\
 & SROCC & 0.891 & 0.883 & 0.844 & 0.872 & 0.631 & 0.619 \\
 & KROCC & 0.685 & 0.691 & 0.663 & 0.670 & 0.462 & 0.442 \\
 & RMSE  & 0.480 & 0.492 & 0.662 & 0.562 & 0.983 & 0.704 \\ \midrule
\multirow{4}{*}{MS-GraphSIM} 
 & PLCC  & 0.945 & 0.892 & 0.877 & \underline{0.918} & 0.616 & 0.673 \\
 & SROCC & 0.893 & 0.890 & 0.865 & \underline{0.898} & 0.623 & 0.662 \\
 & KROCC & 0.693 & 0.701 & 0.680 & \underline{0.709} & 0.459 & 0.479 \\
 & RMSE  & 0.472 & 0.478 & 0.610 & \underline{0.499} & 0.995 & 0.675 \\ \midrule
\multirow{4}{*}{PHM} 
 & PLCC  & 0.950 & \textcolor{bestred}{0.916} & \underline{0.901} & \textcolor{secondblue}{0.941} & 0.772 & \underline{0.696} \\
 & SROCC & \textcolor{secondblue}{0.905} & \textcolor{bestred}{0.909} & \underline{0.895} & \textcolor{secondblue}{0.932} & 0.769 & 0.676 \\
 & KROCC & \underline{0.714} & \textcolor{bestred}{0.735} & \underline{0.716} & \textcolor{secondblue}{0.774} & 0.575 & 0.488 \\
 & RMSE  & \underline{0.452} & \textcolor{bestred}{0.425} & \underline{0.552} & \textcolor{secondblue}{0.425} & 0.803 & \underline{0.656} \\ 
 \midrule
\multirow{4}{*}{TCDM} 
 & PLCC  & 0.943 & \textcolor{secondblue}{0.904} & 0.899 & \textcolor{bestred}{0.952} & 0.821 & 0.689 \\
 & SROCC & 0.891 & \textcolor{secondblue}{0.903} & 0.883 & \textcolor{bestred}{0.939} & 0.818 & \underline{0.678} \\
 & KROCC & 0.697 & \textcolor{secondblue}{0.722} & 0.710 & \textcolor{bestred}{0.782} & \underline{0.622} & \underline{0.493} \\
 & RMSE  & 0.481 & \textcolor{secondblue}{0.454} & 0.558 & \textcolor{bestred}{0.385} & 0.721 & 0.661 \\ 
 \midrule
\multirow{4}{*}{\shortstack[l]{FRSVR\\(K-fold splicing)}} 
 & PLCC  & \underline{0.952} & 0.850 & \textcolor{secondblue}{0.911} & 0.833 & \textcolor{secondblue}{0.890} & \textcolor{secondblue}{0.733} \\
 & SROCC & \underline{0.902} & 0.864 & \textcolor{secondblue}{0.906} & 0.829 & \textcolor{secondblue}{0.879} & \textcolor{secondblue}{0.730} \\
 & KROCC & \textcolor{secondblue}{0.716} & 0.678 & \textcolor{secondblue}{0.733} & 0.638 & \textcolor{secondblue}{0.654} & \textcolor{secondblue}{0.522} \\
 & RMSE  & 0.466 & 0.523 & \textcolor{secondblue}{0.511} & 0.635 & \textcolor{secondblue}{0.652} & \textcolor{secondblue}{0.621} \\ \midrule
\multirow{4}{*}{\shortstack[l]{PointPCA\\(K-fold splicing)}}  
 & PLCC  & 0.942 & 0.847 & 0.879 & 0.820 & \underline{0.842} & 0.655 \\
 & SROCC & 0.891 & 0.859 & 0.885 & 0.819 & \underline{0.823} & 0.671 \\
 & KROCC & 0.703 & 0.679 & 0.702 & 0.634 & 0.611 & 0.481 \\
 & RMSE  & 0.471 & 0.517 & 0.572 & 0.627 & \underline{0.711} & 0.669 \\ \midrule
\multirow{4}{*}{\shortstack[l]{\textbf{MS-ISSM}\\\textbf{(K-fold splicing)}}}  
 & PLCC  & \textcolor{secondblue}{0.954} & 0.876 & \textcolor{bestred}{0.933} & 0.916 & \textcolor{bestred}{0.902} & \textcolor{bestred}{0.783} \\
 & SROCC & 0.890 & 0.862 & \textcolor{bestred}{0.923} & 0.889 & \textcolor{bestred}{0.904} & \textcolor{bestred}{0.754} \\
 & KROCC & 0.695 & 0.689 & \textcolor{bestred}{0.767} & 0.699 & \textcolor{bestred}{0.730} & \textcolor{bestred}{0.562} \\
 & RMSE  & \textcolor{secondblue}{0.432} & 0.512 & \textcolor{bestred}{0.459} & 0.529 & \textcolor{bestred}{0.545} & \textcolor{bestred}{0.568} \\ \bottomrule
\end{tabular}
\end{adjustbox}
\label{table_distortion}
\end{table}

\subsection{Evaluation Criteria}
To ensure alignment between the subjective ratings and objective predictions of different metrics, we standardize the objective predictions to a consistent dynamic range based on guidance from the video quality expert group (VQEG) \cite{ref44}. Subsequently, we use Pearson's linear correlation coefficient (PLCC), Spearman's rank order correlation coefficient (SROCC), Kendall's rank order correlation coefficient (KROCC), and root mean square error (RMSE) to evaluate the performance of various metrics, representing their linearity, monotonicity, and accuracy, respectively. Higher PLCC, SROCC, and KROCC values indicate superior metric performance, while lower RMSE values suggest better accuracy. To normalize the scores of objective quality assessment metrics onto a uniform scale, we apply the logistic regression method recommended by VQEG. \par

\subsection{Datasets and PCQA Metrics} 
To verify the performance of the proposed method across different types of distortions, we use four point cloud subjective datasets for assessment. These datasets include: SJTU \cite{ref14}, WPC \cite{ref39}, M-PCCD\cite{ref42}, and ICIP \cite{ref43}. 
\ADDCZ{Moreover, to optimize the regression model and guarantee strict content separation, we employed a K-fold cross-validation strategy on each dataset\cite{refadd6}. To accommodate the varying sizes of the datasets, the number of folds was explicitly tailored: we utilized 6-fold cross-validation for the ICIP dataset, 8-fold for M-PCCD, 9-fold for SJTU, and 10-fold for WPC. Performance is evaluated using two distinct aggregation methods. The first, denoted as ``K-fold average", computes the PLCC, SROCC, KROCC, and RMSE metrics per fold and averages them. The second, denoted as ``K-fold splicing", concatenates the out-of-sample prediction scores for all point clouds across the K-fold iterations to compute the evaluation metrics globally at once \cite{refadd6}.}
In addition, this paper compares the proposed algorithm with \ADDCZ{12} classic and SOTA PCQA metrics. \ADDCZ{It is worth noting that PointPCA\cite{refadd5} and FRSVR\cite{refadd8} are learning-based methods, and their training and testing follow the same procedures as the proposed method.}\par

\subsection{Performance Comparison}
We evaluate the performance of various PCQA metrics using different datasets. The overall evaluation results for each PCQA metric across these datasets are presented in Table \ref{table_all}. To facilitate direct comparison, \ADDCZ{the best, second-best, and third-best performing metrics are highlighted in red, blue, and underlined text, respectively. Notably, the K-fold average results of the learning-based methods are excluded from the comparison and ranking. Instead, the rankings rely on the global concatenated results using splicing. This approach aggregates predictions from all test folds before calculating the final correlation. It ensures a fair comparison with traditional hand-crafted metrics evaluated on the entire dataset.} Specifically, the proposed method ranks highest on the WPC dataset compared to the other PCQA metrics. On the M-PCCD dataset, \ADDCZ{the proposed method achieves first place in PLCC, SROCC, and KROCC.} On the ICIP dataset, \ADDCZ{it achieves the best performance in PLCC, SROCC, KROCC, and RMSE}. Although the proposed method does not achieve the best performance on the SJTU dataset, \ADDCZ{it closely trails the top-performing TCDM method. On the SJTU dataset, MS-ISSM yields a PLCC of 0.906 and an SROCC of 0.897, compared with the best-performing method, TCDM, which achieves 0.931 and 0.912. Notably, based on the overall rank across all four datasets, the proposed method demonstrates the best performance. It is followed by TCDM, which achieves competitive but lower correlations. Furthermore, compared with other learning-based methods, our method demonstrates better overall performance in both K-fold average and K-fold splicing results. Overall, the proposed method is demonstrated with advanced performance.} \par 

\begin{figure*}[p]
\centering  
\includegraphics[width=5.8in]{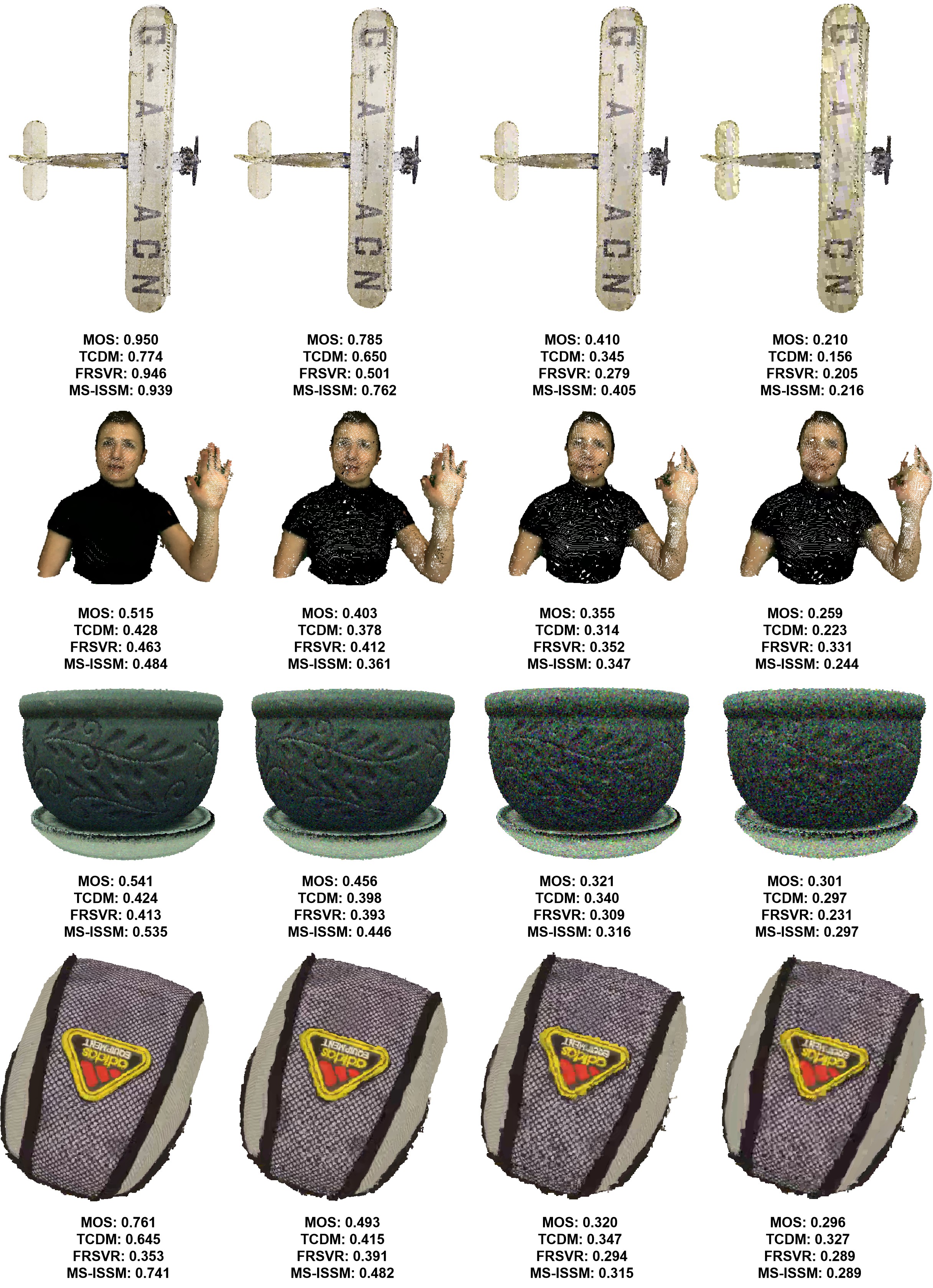}
\caption{\ADDCZ{Qualitative visual comparisons of distorted point clouds and their objective quality scores. From top to bottom, the models are subjected to octree-based compression ($biplane$), trisoup-based compression ($sarah$), Gaussian noise ($flowerpot$), and video compression ($bag$), with distortion severity increasing from left to right. The scores below each model represent the subjective MOS, predictions from two SOTA methods (TCDM and FRSVR), and our proposed MS-ISSM.}}
\label{fig:visual}
\end{figure*}
\ADDCZ{We combine four datasets to validate metric performance across distortion types. Because subjective rating scales differ, we map all scores onto a common scale of 0 to 5 for consistency. Table \ref{table_distortion} presents the results across distortions including octree-based compression (Oc), video-based compression (Vc), trisoup-based compression (Tc), noise (Ns), downsampling (Ds), and mixing (Mix). As with previous tables, the best, second-best, and third-best performing metrics highlighted in red, blue, and underlined text.}
\ADDCZ{Overall, the proposed MS-ISSM exhibits strong performance in octree-based compression, trisoup-based compression, and video-based compression. Specifically, it achieves the highest PLCC, SROCC, KROCC, and RMSE in these three categories. For downsampling distortion, MS-ISSM performs competitively, securing the second-best PLCC and RMSE.} \par
\ADDCZ{Regarding mixing distortion, MS-ISSM performs lower than TCDM. The mixing distortion samples originate entirely from the SJTU dataset, where TCDM parameters were explicitly fitted, providing a distributional advantage. However, in the video-based compression scenario, MS-ISSM maintains a clear lead. While FRSVR obtains an SROCC of 0.730 and an RMSE of 0.621, MS-ISSM reaches a higher SROCC of 0.754 and a lower RMSE of 0.568.} \par
\ADDCZ{Overall, MS-ISSM achieves the strong performance across different distortion types.} 
This generalization capability confirms the effectiveness of our design \ADDCZ{in three key aspects.} \ADDCZ{First,} the RBF-based implicit representation reconstructs continuous surfaces to resolve sparsity and geometric jitter. This bypasses discrete point-to-point matching errors, yielding superior stability in V-PCC and compression distortions. \ADDCZ{Second,} our hierarchical approach captures both global structural shifts from downsampling and local high-frequency artifacts from compression. \ADDCZ{Finally,} the ResGrouped-MLP combines Log-Modulus transformation for distribution rectification with channel-wise attention for adaptive feature weighting, ensuring robust prediction in mixed distortion scenarios.\par
Additionally, to further compare the performance of the MS-ISSM method and the RBFIM method in terms of compression distortion, we evaluated datasets containing compression distortion from the three aforementioned datasets. Based on the type of compression distortion, the datasets were divided into G-PCC compression distortion and V-PCC compression distortion. \ADDCZ{Table \ref{table_compression} shows that RBFIM performs better on both subsets of the ICIP dataset. However, MS-ISSM outperforms RBFIM on the G-PCC and V-PCC subsets of the WPC and M-PCCD datasets. Furthermore, MS-ISSM achieves higher correlations and lower errors on the combined ALL dataset. This indicates that the multi-scale implicit feature method provides stable generalization across diverse compression scenarios.} \par
\begin{table}[h]
\centering
\setlength{\tabcolsep}{5.5pt}
\caption{\ADDCZ{Comparison of the RBFIM and MS-ISSM (K-fold splicing) methods under different compression distortions.}}
\begin{tabular}{cccc}
\hline
Datasets                    & criteria & {RBFIM\cite{ref20}} & MS-ISSM \\ \hline
\multirow{4}{*}{ICIP-GPCC}  & PLCC     & 0.993 & 0.975   \\
                            & SROCC    & 0.971 & 0.968   \\
                            & KROCC    & 0.870 & 0.846   \\
                            & RMSE     & 0.020 & 0.027   \\ \hline
\multirow{4}{*}{ICIP-VPCC}  & PLCC     & 0.969 & 0.969   \\
                            & SROCC    & 0.976 & 0.956   \\
                            & KROCC    & 0.895 & 0.840   \\
                            & RMSE     & 0.067 & 0.130   \\ \hline
\multirow{4}{*}{WPC-GPCC}   & PLCC     & 0.841 & 0.865   \\
                            & SROCC    & 0.847 & 0.864   \\
                            & KROCC    & 0.655 & 0.677   \\
                            & RMSE     & 0.193 & 0.165   \\ \hline
\multirow{4}{*}{WPC-VPCC}   & PLCC     & 0.482 & 0.683   \\
                            & SROCC    & 0.473 & 0.638   \\
                            & KROCC    & 0.343 & 0.461   \\
                            & RMSE     & 0.190 & 0.176   \\ \hline
\multirow{4}{*}{MPCCD-GPCC} & PLCC     & 0.734 & 0.977  \\
                            & SROCC    & 0.673 & 0.974   \\
                            & KROCC    & 0.544 & 0.866   \\
                            & RMSE     & 0.280 & 0.096   \\ \hline
\multirow{4}{*}{MPCCD-VPCC} & PLCC     & 0.489 & 0.921   \\
                            & SROCC    & 0.460 & 0.904   \\
                            & KROCC    & 0.288 & 0.750   \\
                            & RMSE     & 0.245 & 0.117   \\ \hline
\multirow{4}{*}{ALL}        & PLCC     & 0.611 & 0.892   \\
                            & SROCC    & 0.566 & 0.892   \\
                            & KROCC    & 0.406 & 0.718   \\
                            & RMSE     & 0.259 & 0.159   \\ \hline
\end{tabular}
\label{table_compression}
\end{table}
\subsection{\ADDCZ{Visual Comparisons}}
\ADDCZ{To illustrate the perceptual performance of MS-ISSM against existing methods, we present a qualitative comparison in Fig. \ref{fig:visual}.  We selected four representative point clouds subjected to distinct distortion types: $biplane$ with octree-based compression, $sarah$ with trisoup-based compression, $flowerpot$ with Gaussian noise, and $bag$ with video-based compression. We then compared the subjective MOS with the predicted scores from the multi-scale feature-based TCDM, the learning-based FRSVR, and our proposed MS-ISSM.}\par
\ADDCZ{The visual analysis shows that MS-ISSM accurately captures steady quality drops as distortion levels increase. For example, in $biplane$ with octree-based compression, our method tracks the MOS drop from 0.950 to 0.210 by predicting scores from 0.939 to 0.216, which avoids the consistent underestimation of TCDM and the excessive penalties of FRSVR at medium distortion levels. Furthermore, MS-ISSM is robust to noise, as seen in $flowerpot$ with Gaussian noise where TCDM and FRSVR give excessively low scores to small point shifts. In contrast, MS-ISSM uses continuous RBF functions to smooth out these small errors, staying consistent with the HVS. Additionally, in $bag$ with video-based compression, FRSVR incorrectly gives a low score of 0.353 to a high-quality model with an MOS of 0.761, causing an incorrect quality ranking. Meanwhile, MS-ISSM correctly tracks the actual quality drop from 0.741 to 0.289. Overall, this visual analysis confirms that MS-ISSM aligns well with human visual perception.}\par

\subsection{\ADDCZ{Sensitivity Analysis}}
\ADDCZ{We conducted a sensitivity analysis using the color noise (CN) and geometry noise (GN) subsets from the SJTU dataset to evaluate whether the differences in implicit function coefficients well reflect perceptual distortion. The SJTU dataset defines CN as photometric noise applied to the RGB attributes of randomly selected points. Specifically, the noise is injected into 10\%, 30\%, 40\%, 50\%, 60\%, and 70\% of the points, with corresponding intensity limits set to 10, 30, 40, 50, 60, and 70. GN applies a random Gaussian distributed geometric shift to all points. These shifts are bounded within 0.05\%, 0.1\%, 0.2\%, 0.5\%, 0.7\%, and 1.2\% of the bounding box. We computed  the difference in the implicit function coefficient (45-dimensional) for the original and distorted point cloud respectively, and evaluated their correlation in PLCC with the perceptual quality measured in subjective MOS. The results are then compared with traditional point-to-point metrics (PSNR-p2p and PSNR-YUV) on the above subsets. }\par
\begin{figure}[]
   \centering
   \subfloat[CN]{\label{fig:i}\includegraphics[width=2.60in]{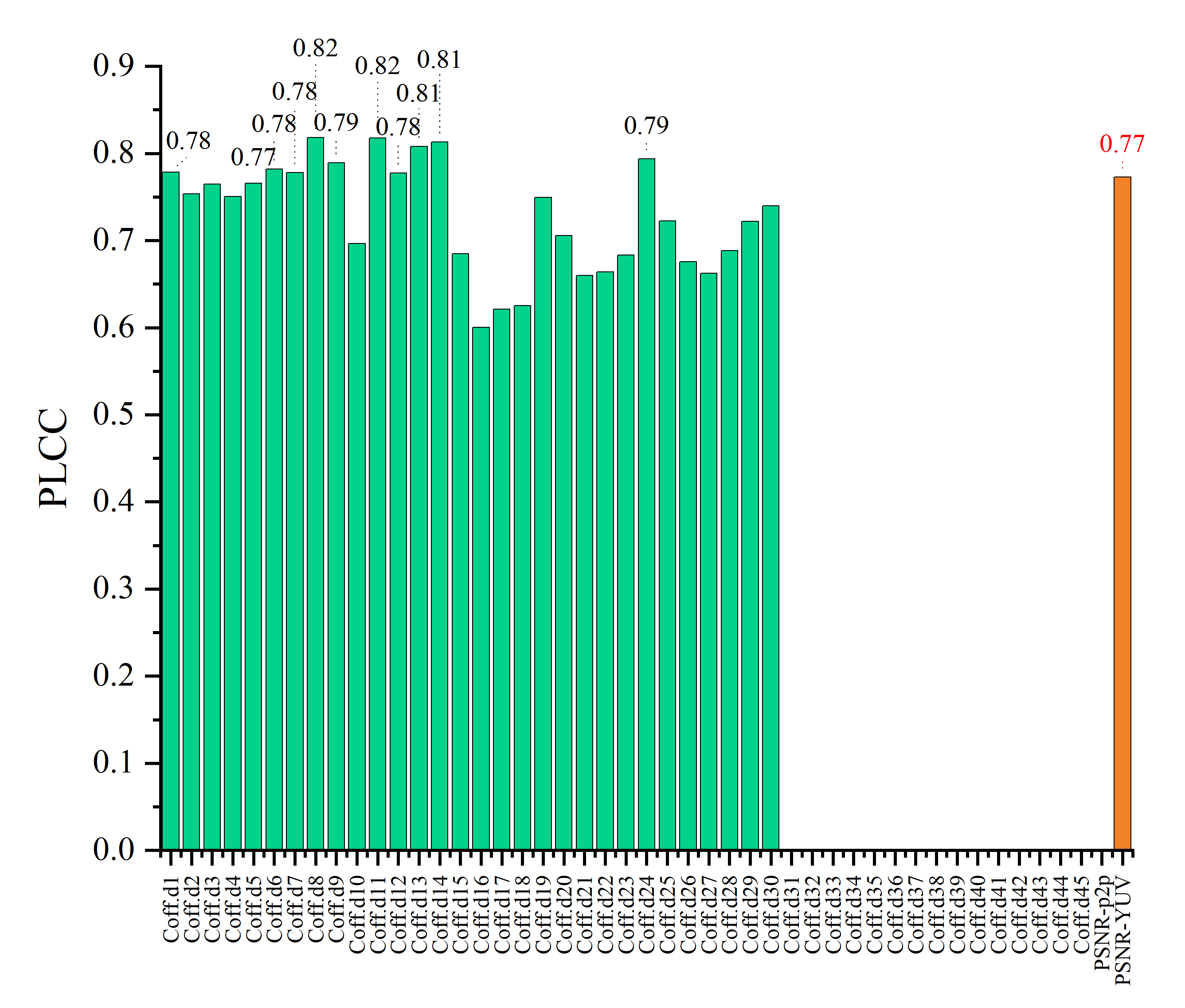}}\\
   \subfloat[GN]{\label{fig:e}\includegraphics[width=2.55in]{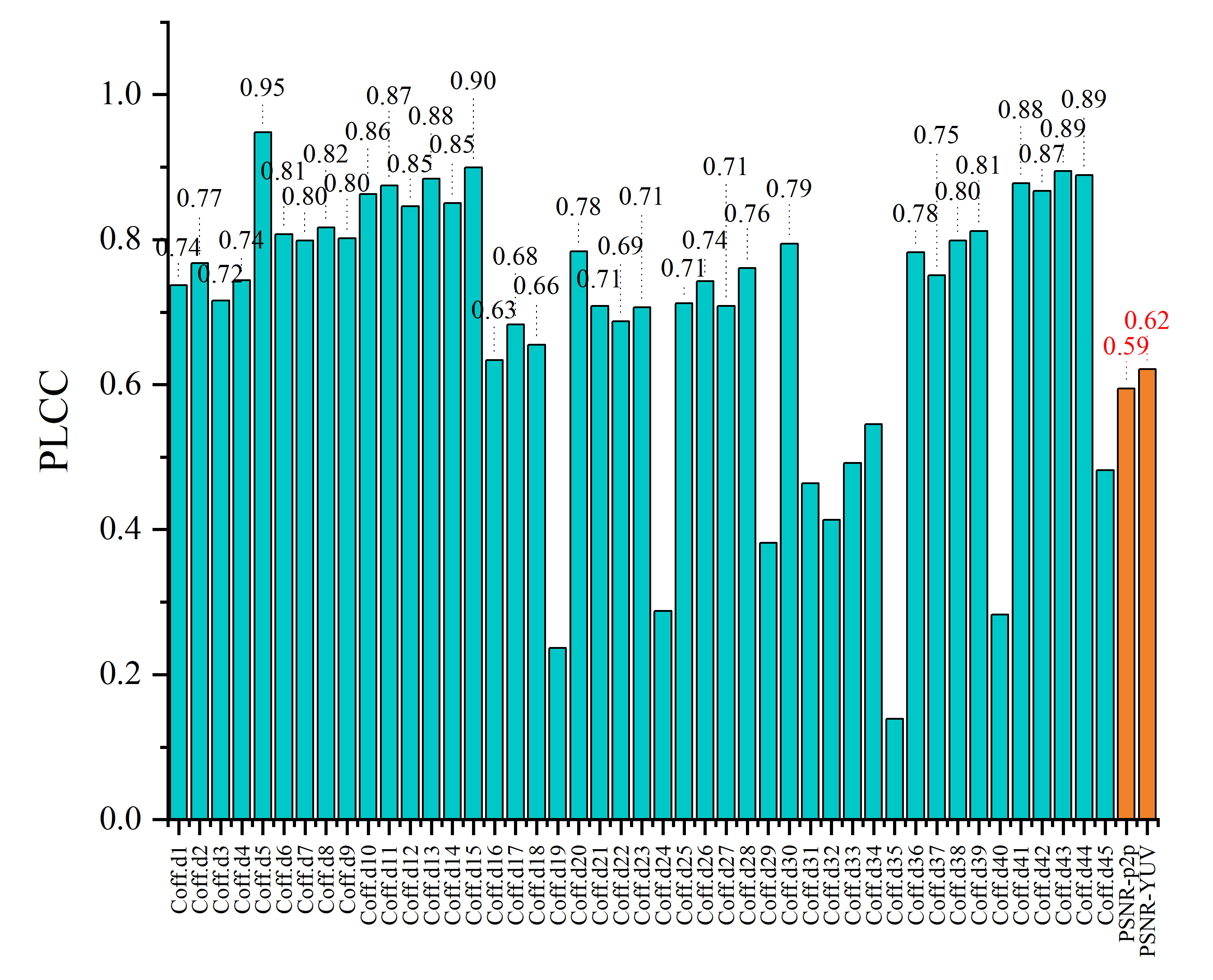}}\\
   \caption{\ADDCZ{Sensitivity analysis of the RBF coefficient differences.}}
   \label{sensitivity}
  \end{figure}
\ADDCZ{As shown in Fig. \ref{sensitivity}, the quantitative results demonstrate that the RBF coefficient differences exhibit a strong correlation with MOS not only at their peak values but across the broader 45-dimensional feature distribution. Under CN, our features achieve a maximum correlation of 0.82, with 11 out of the 30 color-related dimensions explicitly outperforming the traditional PSNR-YUV metric at 0.77, while PSNR-p2p is inherently inapplicable to pure color distortions. Under GN, the coefficients demonstrate even stronger sensitivity with a maximum correlation of 0.95. Furthermore, this superiority is widely distributed across the feature space under GN, where 8 out of the 15 geometry-related dimensions surpass PSNR-p2p at 0.59, and 27 out of the 30 color-related dimensions exceed PSNR-YUV at 0.62. These findings empirically validate that implicit coefficients capture perceptual degradation much more reliably and comprehensively than discrete point matching. }\par
\begin{figure}[h]
\centering  
\includegraphics[width=3.0in]{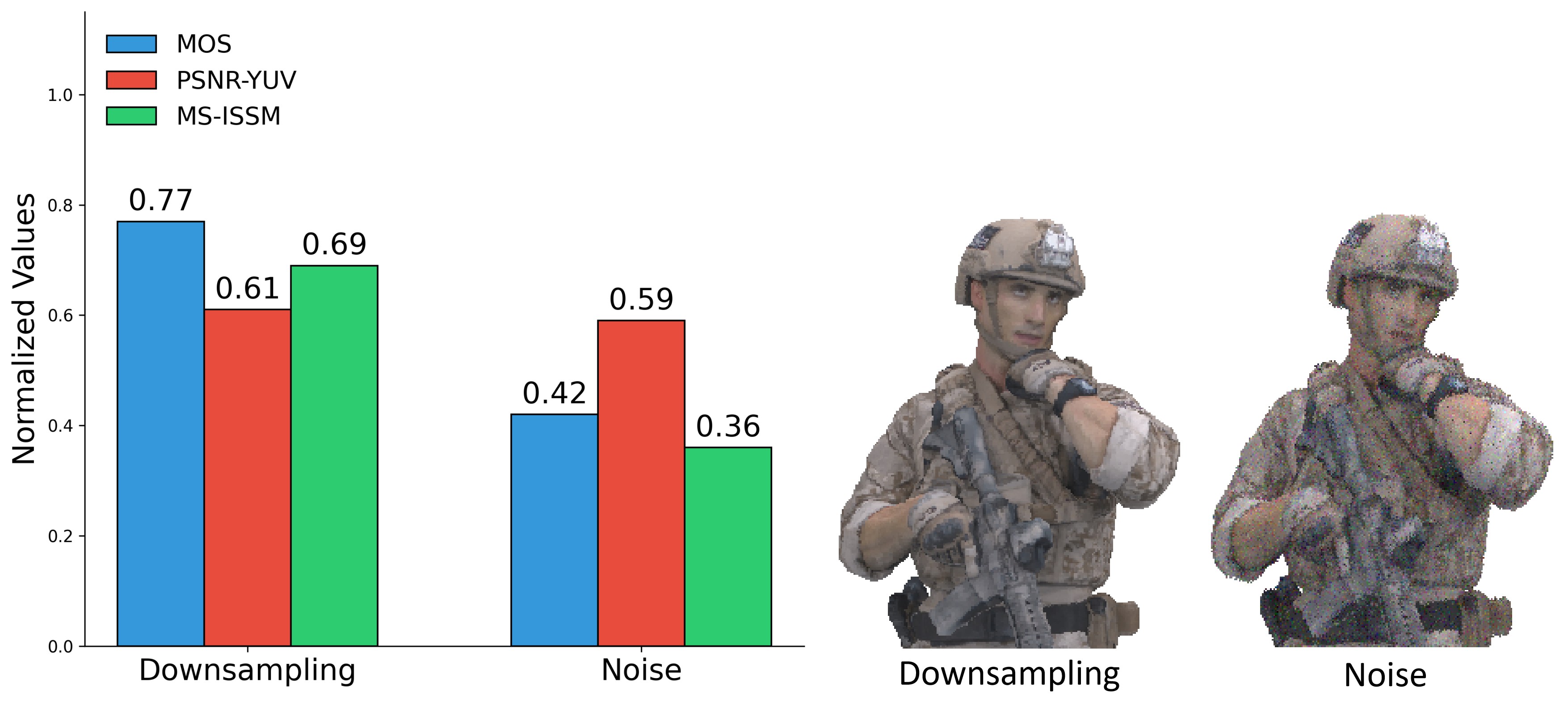}
\caption{\ADDCZ{Coefficient-space difference vs. Point-space error.}}
\label{compared}
\end{figure}

\ADDCZ{Moreover, to better interpret the superiority of coefficient-space difference, we analyzed $soldier$ subjected to two different types of distortions: noise and downsampling.} 
\ADDCZ{As shown in Fig. \ref{compared}, we compare the predicted normalized quality scores of PSNR-YUV and our MS-ISSM against the subjective MOS. Because the unstructured nature of point clouds leads to local point coordinate shifts in both scenarios, the traditional point-to-point metric (PSNR-YUV) fails to distinguish their actual perceptual differences, producing almost identical quality scores of 0.61 and 0.59 for downsampling and noise, respectively. In contrast, human vision is relatively tolerant to uniform downsampling with a high MOS of 0.77, but highly sensitive to noise-induced degradation with a low MOS of 0.42. Our MS-ISSM effectively captures this perceptual discrepancy, predicting a high quality score of 0.69 for downsampling and appropriately penalizing the noise distortion with a low score of 0.36. Ultimately, the quality predictions of MS-ISSM align consistently with human visual perception. }\par

\subsection{\ADDCZ{Complexity Analysis}}
\begin{figure}[h]
\centering  
\includegraphics[width=2.5in]{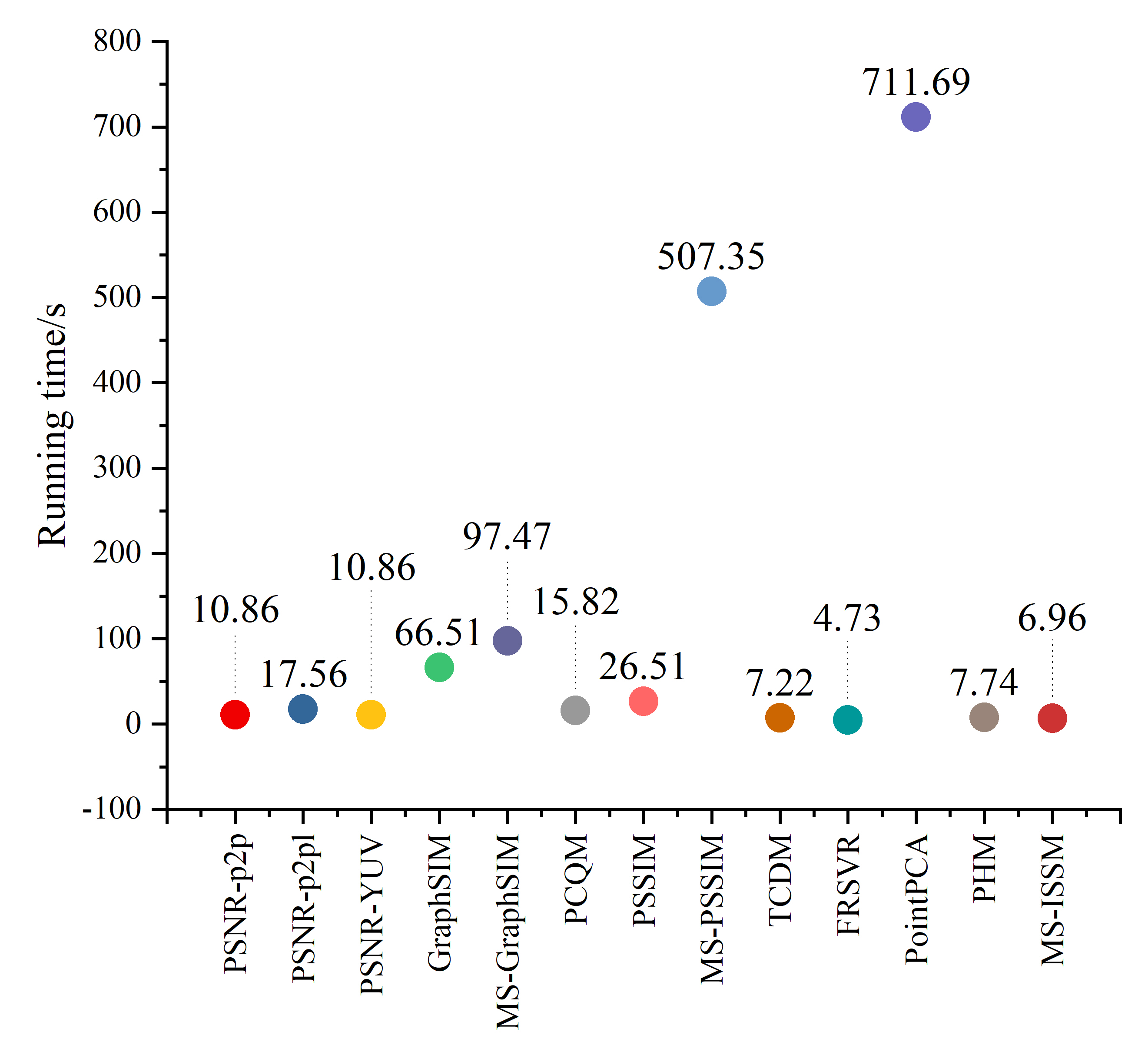}
\caption{\ADDCZ{Time complexity of PCQA methods on four datasets.}}
\label{runningtime}
\end{figure}

We compare the average running times across four datasets on an Intel Core i7-8809G CPU @3.10GHz. As shown in Fig. \ref{runningtime}, our method achieves superior efficiency, surpassed only by FRSVR. \ADDCZ{The efficiency of MS-ISSM comes from replacing global topological comparisons with local matrix operations. When the original point cloud has $N_{\rm{O}}$ points, the sampled reference point set has $N_{\rm{R}}$ points where $N_{\rm{R}} \ll N_{\rm{O}}$, and the local neighborhood size is set to 30. The algorithm processes the data in several steps. First, iterating points to build multi-scale voxel representations takes $\mathcal{O}(N_{\rm{O}})$ time. Next, building a KD-Tree for spatial partitioning requires $\mathcal{O}(N_{\rm{O}} \log N_{\rm{O}})$ operations, and finding nearest neighbors for the $N_{\rm{R}}$ reference points takes $\mathcal{O}(N_{\rm{R}} \log N_{\rm{O}})$ time. For the extraction of curvature features, calculating the local surface variations involves secondary constant-sized neighborhood searches and matrix decompositions within each patch, which securely bounds to an additional ${\mathcal O}({N_{\rm{R}}}\log {N_{\rm{O}}})$ time. Subsequently, solving constant-sized linear systems for local RBF fitting adds $\mathcal{O}(N_{\rm{R}})$ time, while processing feature vectors via the ResGrouped-MLP takes $\mathcal{O}(1)$ time. As summarized in Table \ref{tab:complexity}, the overall complexity is dominated by space partitioning and neighbor search. Since these operations are performed on both the original and distorted point clouds, the total time doubles but remains bounded by $\mathcal{O}(N_{\rm{O}} \log N_{\rm{O}})$. Graph-based metrics such as GraphSIM and MS-GraphSIM require graph construction or spectral decomposition, while MS-PSSIM and PointPCA are burdened by high-dimensional multi-scale processing in the point space. In contrast, MS-ISSM avoids global topology reconstruction; by restricting complex matrix calculations to highly downsampled reference points, it ensures the linearithmic efficiency shown in Fig. \ref{runningtime}.} Although simple single-scale metrics like p2p remain computationally light, our method offers a better trade-off between processing speed and multi-scale performance, making it highly suitable for large-scale PCQA.\par

\begin{table}[htbp]
\centering
\caption{\ADDCZ{complexity analysis of the proposed MS-ISSM.}}
\label{tab:complexity}
\begin{adjustbox}{max width=\linewidth}
\begin{tabular}{l c}
\toprule
{Processing Stage} & {Complexity} \\ 
\midrule
\begin{tabular}{@{}l@{}}multi-scale voxelization\end{tabular} & $\mathcal{O}(N_{\rm{O}})$ \\ 
\addlinespace
\begin{tabular}{@{}l@{}}curvature extraction\end{tabular} & ${\mathcal O}({N_{\rm{R}}}\log {N_{\rm{O}}})$ \\ 
\addlinespace
\begin{tabular}{@{}l@{}}space partitioning\end{tabular} & $\mathcal{O}(N_{\rm{O}} \log N_{\rm{O}})$ \\ 
\addlinespace
\begin{tabular}{@{}l@{}}KNN search\end{tabular} & $\mathcal{O}(N_{\rm{R}} \log N_{\rm{O}})$ \\ 
\addlinespace
\begin{tabular}{@{}l@{}}RBF implicit fitting\end{tabular} & $\mathcal{O}(N_{\rm{R}})$ \\ 
\addlinespace
\begin{tabular}{@{}l@{}}quality regression\end{tabular} & $\mathcal{O}(1)$ \\ 
\midrule
\underline{overall complexity} & \textbf{$\mathcal{O}(N_{\rm{O}} \log N_{\rm{O}})$} \\ 
\bottomrule
\end{tabular}
\end{adjustbox}
\end{table}

\begin{figure}[h]
\centering  
\includegraphics[width=2.2in]{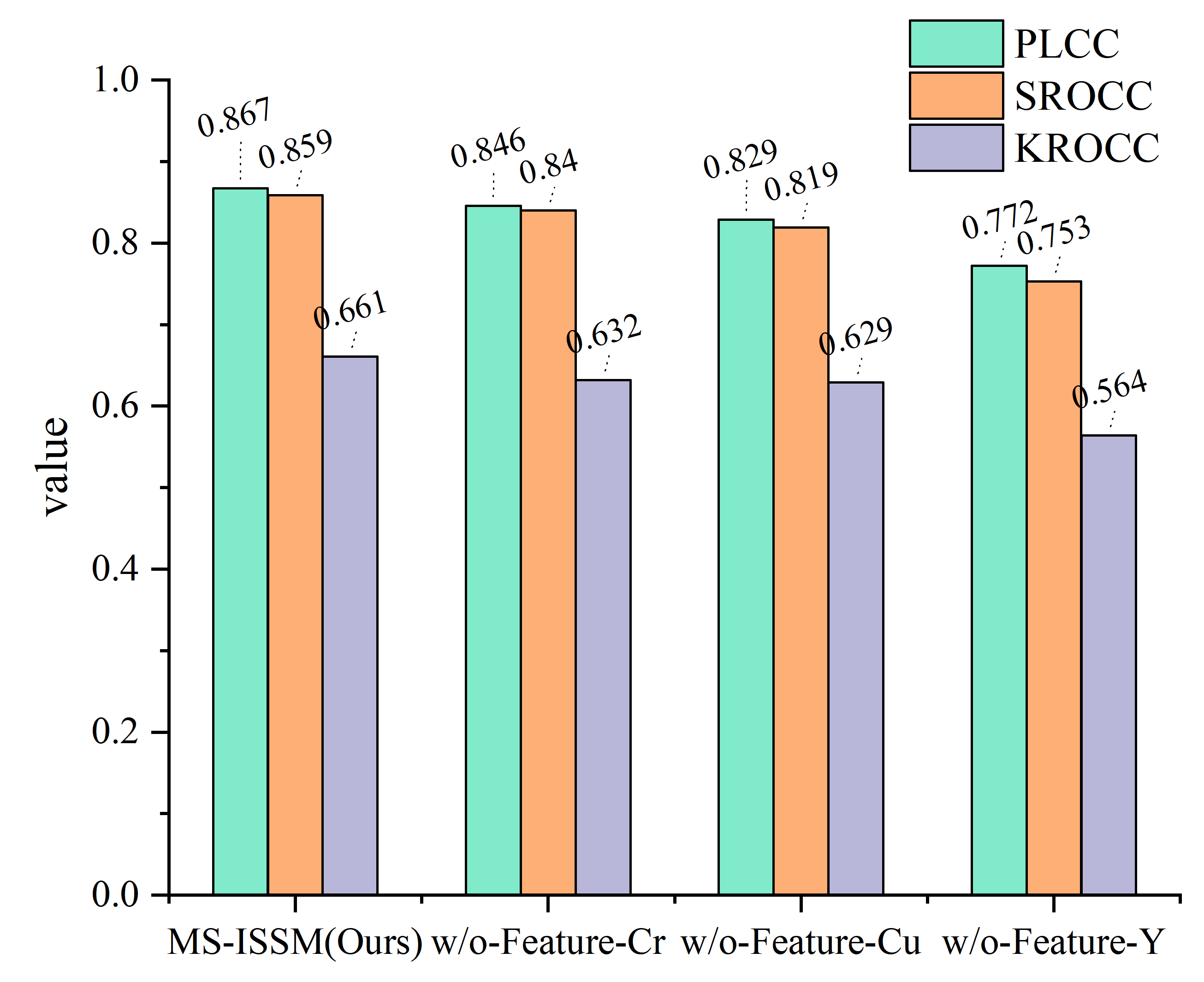}
\caption{\ADDCZ{Performance comparison among different feature types.}}
\label{fig6}
\end{figure}

\subsection{Ablation Studies}
To comprehensively validate the effectiveness of the proposed MS-ISSM framework, we conducted extensive ablation studies. These experiments are designed to investigate the contribution of \ADDCZ{four} key aspects: the multi-modal implicit features, the multi-scale strategy, the specific architectural components of the ResGrouped-MLP network, \ADDCZ{and the key hyperparameters of MS-ISSM}. The results are analyzed on the  \ADDCZ{WPC} dataset to ensure statistical reliability.
\begin{itemize}
\item{Impact of Multi-modal Implicit Features}
\end{itemize}
We first investigate the impact of different basis feature functions by evaluating the performance of MS-ISSM \ADDCZ{when removing single feature types: w/o luma, chroma, or curvature}. As illustrated in Fig. \ref{fig6}, \ADDCZ{while the incomplete models can still provide reasonable quality estimations, omitting any specific feature, particularly luma, exhibits limitations in capturing the full spectrum of perceptual distortions.} For instance, geometry distortions are less perceptible when structural features are ablated, and vice versa for color. The complete fusion of multi-modal features yields superior robustness, confirming that combining geometry and color attributes is essential for aligning with the HVS. \par
\begin{itemize}
\item{Impact of Multi-scale Strategy}
\end{itemize}
To verify the necessity of the multi-scale hierarchy, we evaluated the performance \ADDCZ{by ablating individual spatial scales: w/o High, Medium, or Low}. As reported in Table \ref{table7}, although \ADDCZ{models missing a single scale still} achieve decent correlations, their performance fluctuates across different datasets due to varying point cloud densities and content characteristics. For example, \ADDCZ{omitting} the High scale \ADDCZ{results in a more significant performance drop} on datasets with fine textures, while \ADDCZ{removing} the Low scale more \ADDCZ{severely impairs the evaluation of} global structural distortions. By integrating all three scales, MS-ISSM effectively aggregates local details and global topology, achieving consistent and optimal performance across diverse datasets.\par
\begin{table}[h]
\centering
\setlength{\tabcolsep}{4.8pt}
\caption{\ADDCZ{The Performance comparison of different scales.}}
\begin{adjustbox}{max width=\textwidth}
\begin{tabular}{cccccc}
\toprule
Dataset & Scale & PLCC & SROCC & KROCC & RMSE \\ \midrule
\multirow{4}{*}{WPC} & MS-ISSM (Ours) & 0.867 & 0.859 & 0.661 & 11.618 \\
 & w/o-Scale-H & 0.846 & 0.832 & 0.642 & 12.227 \\
 & w/o-Scale-L & 0.833 & 0.827 & 0.636 & 12.674 \\
 & w/o-Scale-M & 0.849 & 0.839 & 0.650 & 12.098 \\
 \bottomrule
\end{tabular}
\end{adjustbox}
\label{table7}
\end{table}

\begin{itemize}
\item{Impact of ResGrouped-MLP Architecture}
\end{itemize}
Finally, to justify the design rationale of our regression network, we conducted an ablation study by removing or replacing key modules in the ResGrouped-MLP. The comparison results are summarized in Table \ref{ablation}. 
\ADDCZ{To evaluate the impact of the Log-Modulus transformation,} we removed the Log-Modulus preprocessing and used standard Z-score normalization directly on the raw coefficients. As shown in Table \ref{ablation}, this resulted in a performance drop, with SROCC decreasing by \ADDCZ{0.015}. This confirms that the raw statistical features follow a heavy-tailed distribution, and the proposed Log-Modulus transformation effectively suppresses outliers, enabling the network to learn more robust feature representations. 
To validate the ``Split-Transform-Merge" strategy, we replaced the grouped encoders with a standard MLP that concatenates all multi-scale features at the input stage. The results show that the grouped strategy outperforms early concatenation by \ADDCZ{0.007} in PLCC. This suggests that processing luma, chroma, and curvature features independently in the early layers prevents information interference, allowing the network to capture distinct distortion patterns for each channel. We also removed the scale-wise channel attention blocks \ADDCZ{to assess their contribution}. The \ADDCZ{subsequent} decline in performance indicates that the attention mechanism plays a crucial role \ADDCZ{in allowing} the model to adaptively recalibrate the importance of different channels, thereby better mimicking the varying sensitivities of the HVS. 
In summary, each component of the MS-ISSM, from implicit feature extraction to the hierarchical regression network, makes a significant contribution to the final prediction accuracy and robustness. \par

\begin{table}[]
\centering
\setlength{\tabcolsep}{4.8pt}
 \caption{\ADDCZ{ABLATION STUDY OF THE PROPOSED RESGROUPED-MLP ARCHITECTURE ON THE WPC DATASET.}}
  \begin{adjustbox}{max width=\textwidth}
\begin{tabular}{ccccc}
\hline
Model Variant        & PLCC  & SROCC & KROCC & RMSE  \\ \hline
MS-ISSM (Ours)       & 0.867 & 0.859 & 0.661 & 11.618 \\
w/o Log-Modulus      & 0.855 & 0.844 & 0.657 & 11.878 \\
w/o Grouped Encoders & 0.860 & 0.848 & 0.662 &  11.691\\
w/o Attention Block  & 0.852 & 0.841 & 0.652 & 11.987 \\ \hline
\end{tabular}
\end{adjustbox}
\label{ablation}
\end{table}

\begin{itemize}
\item{\ADDCZ{Impact of Key Hyperparameters}}
\end{itemize}
\ADDCZ{We conducted experiments to evaluate the impact of the number of RBF neighbors, multi-scale voxel sizes, the attention bottleneck ratio, and the reference points sampling size. We evaluated these variants on the WPC dataset to ensure reliability, with results summarized in Table \ref{ablation}. The underlined values denote our chosen configurations.}

\begin{table}[]
\centering
\setlength{\tabcolsep}{2.0pt}
 \caption{\ADDCZ{Hyperparameter Ablation on WPC Dataset. The underlined values denote our chosen configurations.}}
  \begin{adjustbox}{max width=\textwidth}
\begin{tabular}{l c c c}
\toprule
{Hyperparameter} & {Configuration} & {PLCC} & {SROCC} \\ 
\midrule
\multirow{3}{*}{RBF Neighbors} 
 & 15 & 0.839 & 0.837 \\
 & \underline{30} & 0.867 & 0.859  \\
 & 50 & 0.832 & 0.825  \\ 
\midrule
\multirow{3}{*}{Voxel Sizes} 
 & (1.0, 2.0, 4.0) & 0.849 & 0.843 \\
 & {\underline{(2.0, 4.0, 8.0)}} & 0.867 & 0.859  \\
 & (4.0, 8.0, 16.0) & 0.819 & 0.822 \\ 
\midrule
\multirow{3}{*}{Bottleneck Ratio} 
 & 2 & 0.853 & 0.841 \\
 & \underline{4} & 0.867 & 0.859 \\
 & 8 & 0.846 & 0.840 \\ 
\midrule
\multirow{3}{*}{Reference Point Size} 
 & 64 & 0.833 & 0.828 \\
 & \underline{32} & 0.867 & 0.859 \\
 & 8 & 0.869 & 0.860  \\ 
\bottomrule
\end{tabular}
\end{adjustbox}
\label{ablation}
\end{table}

\ADDCZ{Regarding the RBF neighbors, we tested values of 15, 30, and 50. A smaller neighborhood of 15 yields a localized receptive field sensitive to isolated noise, while a larger neighborhood of 50 results in over-smoothing that masks fine geometry artifacts and increases computational overhead. Setting the value to 30 balances noise robustness and structural sensitivity. For the multi-scale voxel sizes (H, M, L), we evaluated three different progression ranges: a finer scale of (1.0, 2.0, 4.0), our baseline of (2.0, 4.0, 8.0), and a coarser scale of (4.0, 8.0, 16.0). (1.0, 2.0, 4.0) highlights small details but miss large structural changes. Conversely, (4.0, 8.0, 16.0) downsamples the data too much, losing fine artifacts like color quantization noise. }\par
\ADDCZ{When analyzing the channel reduction ratio within our attention blocks, we tested values of 2, 4, and 8. A low ratio of 2 retains redundant channel information and increases the risk of overfitting. A high ratio of 8 over-compresses the latent space, causing a loss of semantic interactions between physical attributes including luma, chroma, and curvature.}\par
\ADDCZ{Finally, we evaluated reference point sampling sizes of 64, 32, and 8. A larger size of 64 leads to insufficient local feature extraction and a noticeable performance drop. A smaller size of 8 yields a marginal performance improvement, but substantially increases the computational burden. Selecting a size of 32 provides the trade-off between computational efficiency and evaluation accuracy. Overall, these empirical validations prove the effectiveness of the architectural choices in MS-ISSM.}

\subsection{\ADDCZ{Generalization Analysis}}
\ADDCZ{To validate the generalization capabilities of our methods, we conducted cross-dataset evaluations using three large-scale benchmark datasets. Specifically, we trained our model entirely on the WPC dataset and tested it directly on the ICIP, SJTU and M-PCCD datasets. To strictly prevent data leakage, the zero-mean normalization mapping fitted on the training set was directly applied to the testing sets. Table \ref{cross-check} below summarizes the results, where SROCC and PLCC values above 0.82 across distinct data distributions can be observed. It is therefore verified that ResGrouped-MLP learns robust and generalized mappings.} \par
\begin{table}[htbp]
\centering
\setlength{\tabcolsep}{3pt}
\caption{\ADDCZ{Cross-dataset evaluation results trained on the WPC dataset.}}
\label{tab:cross_dataset}
\begin{tabular}{l l c c c c}
\toprule
Training & Testing & PLCC & SROCC & KROCC & RMSE \\
\midrule
WPC & SJTU & 0.840 & 0.827 & 0.634 & 1.298 \\
WPC & M-PCCD & 0.895 & 0.918 & 0.658 & 0.560 \\
WPC & ICIP & 0.914 & 0.918 & 0.753 & 0.497 \\
\bottomrule
\label{cross-check}
\end{tabular}
\end{table}

\section{Conclusion}\label{sec7}
This paper presents a multi-scale implicit structural similarity (MS-ISSM) method for point cloud quality assessment (PCQA). To avoid the accumulation of matching errors in unstructured point clouds, the method leverages implicit functions to represent multi-scale features and evaluates quality based on differences in their coefficients. A ResGrouped-MLP network is introduced, incorporating a Log-Modulus transformation that stabilizes gradient descent and accelerates convergence. The architecture employs a grouped encoding strategy combined with residual blocks and channel-wise attention, enabling the model to preserve distinct physical semantics of luma, chroma, and geometry while adaptively highlighting the most salient distortions across high, medium, and low scales. Experiments demonstrate that MS-ISSM outperforms existing PCQA metrics on public datasets, providing a reliable and consistent quality evaluation. \par

\begin{IEEEbiography}
[{\includegraphics[width=1in,height=1.2in,clip,keepaspectratio]{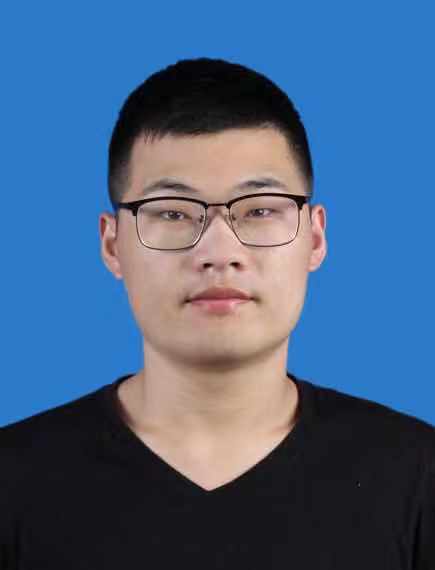}}]{Zhang Chen} received the B.E. degree in 
Electronics and Information Engineering and the M.E. degree in Signal and Information Processing from Northwestern Polytechnical University, Xi’an, China, in 2019 and 2022, respectively. He is currently pursuing the Ph.D. degree in Information and Communication Engineering at the same university. His current research interests include point cloud quality assessment, point cloud compression, and 3-D reconstruction.

\end{IEEEbiography}

\begin{IEEEbiography}
[{\includegraphics[width=1in,height=1.2in,clip,keepaspectratio]{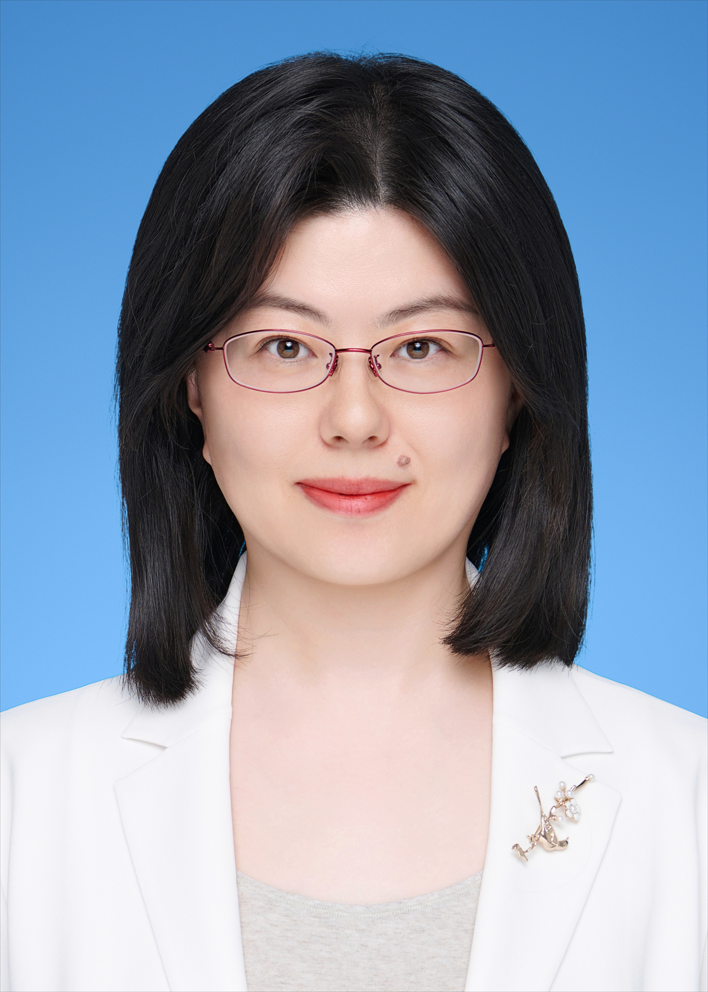}}]{Shuai Wan} (Member, IEEE) received the B.E. degree in Telecommunication Engineering and the M.E. degree in Communication and Information System from Xidian University, Xi’an, China, in 2001 and 2004, respectively, and obtained the Ph.D. in Electronic Engineering from Queen Mary, University of London in 2007. She is currently a Professor at Northwestern Polytechnical University in Xi’an, China. Previously, she served as a Professor at the Royal Melbourne Institute of Technology in Melbourne, Australia, from 2016 to 2025. Her research interests include scalable/multiview video coding, video quality assessment and hyperspectral image compression.
\end{IEEEbiography}

\begin{IEEEbiography}
[{\includegraphics[width=1in,height=1.2in,clip,keepaspectratio]{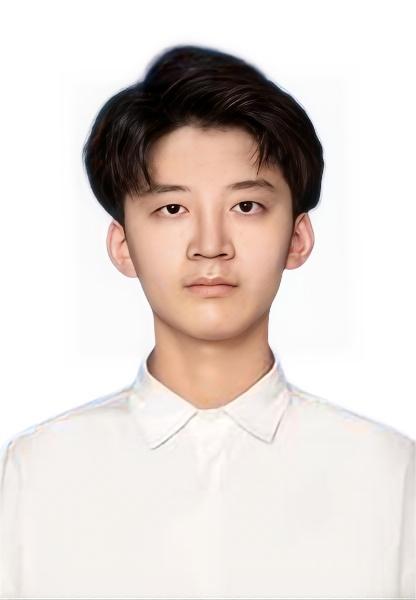}}]{Yuezhe Zhang} received the B.S. degree in electronic information engineering from the School of Electronics and Information, Northwestern Polytechnical University, Xi’an, China, in 2025, where he is currently pursuing the M.S. degree in information and communication engineering. His research interests include 3D Gaussian Splatting: 3DGS data acquisition, registration and alignment, editing, and generation.
\end{IEEEbiography}

\begin{IEEEbiography}
[{\includegraphics[width=1in,height=1.2in,clip,keepaspectratio]{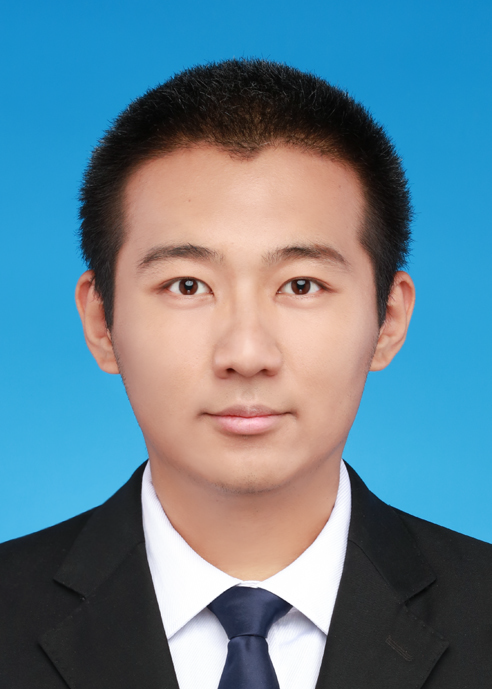}}]{Siyu Ren} received the B.E. degree in Optoelectronic Information Science and Engineering from Tianjin University, Tianjin, China, in 2018. He is currently pursuing the Ph.D. degree in Computer Science at the City University of Hong Kong and Optical Engineering at Tianjin University. His research interests include deep learning and 3D point cloud processing.
\end{IEEEbiography}

\begin{IEEEbiography}
[{\includegraphics[width=1in,height=1.2in,clip,keepaspectratio]{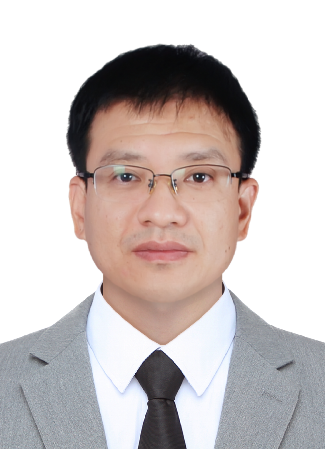}}]{Fuzheng Yang}(Member, IEEE) received the B.E. degree in Telecommunication Engineering, the M.E. degree and the Ph.D. in Communication and Information System from Xidian University, Xi’an, China, in 2000, 2003 and 2005, respectively. He became a lecturer and an Associate Professor in Xidian University in 2005 and 2006, respectively. He has been a professor of communications engineering with Xidian University since 2012. He is also an Adjunct Professor of School of Engineering in RMIT University. During 2006-2007, he served as a visiting scholar and postdoctoral researcher in Department of Electronic Engineering in Queen Mary, University of London. His research interests include video quality assessment, video coding and multimedia communication.
\end{IEEEbiography}

\begin{IEEEbiography}
[{\includegraphics[width=0.95in,height=1.15in,clip,keepaspectratio]{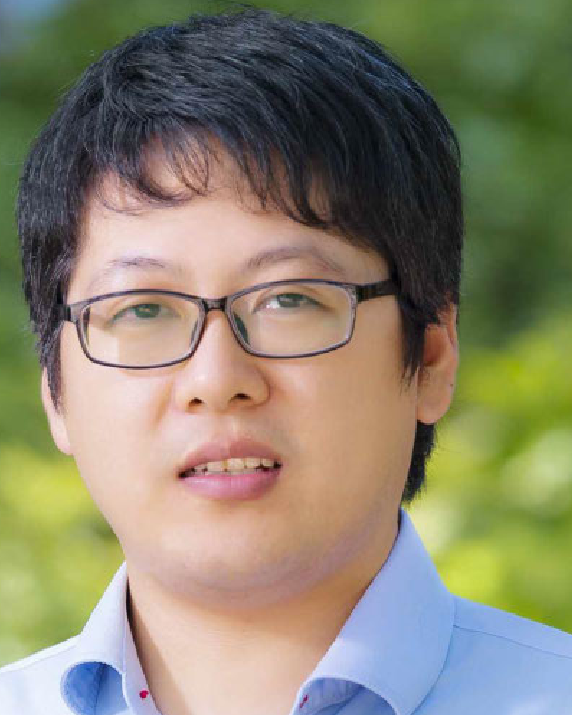}}]{Junhui Hou}(Senior Member, IEEE) is a Professor with the Department of Computer Science, City University of Hong Kong His research interests include multidimensional visual computing, such as light field, hyperspectral, geometry, and event data. He received the Early Career Award from the Hong Kong Research Grants Council in 2018, IEEE Multimedia Rising Star Award in 2023, the Excellent Young Scientists Fund from NSFC in 2024, and the IEEE TIP Best Paper Award in 2025. He is serving as a Senior Area Editor for IEEE TIP and an Associate Editor for IEEE TVCG and TMM. He served as an Associate Editor for IEEE TIP and TCSVT.
\end{IEEEbiography}
\end{document}